\documentclass[11pt]{article}

\usepackage[preprint]{acl}

\usepackage{times}
\usepackage{latexsym}

\usepackage[T1]{fontenc}
\usepackage[utf8]{inputenc}

\usepackage{microtype}

\usepackage{inconsolata}

\usepackage{graphicx}

\usepackage{booktabs}
\usepackage{multirow}
\usepackage{amssymb}
\usepackage{algorithm}
\usepackage{algpseudocode}

\usepackage{tabularx}
\usepackage{array}
\usepackage{makecell}
\usepackage{pifont}

\colorlet{mutedred}{red!70!black}
\definecolor{eocvstatus}{HTML}{0F5F7D}
\colorlet{casebaseline}{mutedred}
\colorlet{casewithheld}{gray}
\colorlet{caseeocv}{eocvstatus}

\newcolumntype{Y}{>{\raggedright\arraybackslash}X}

\newcolumntype{I}{>{\centering\arraybackslash}m{0.08\textwidth}}

\newcolumntype{A}{>{\raggedright\arraybackslash}m{0.15\textwidth}}

\newcolumntype{E}{>{\raggedright\arraybackslash}p{0.16\textwidth}}

\usepackage[most]{tcolorbox}
\tcbset{colback=gray!5!white,colframe=gray!40!white,boxrule=0.4pt,arc=1mm,left=2mm,right=2mm,top=1mm,bottom=1mm}

\newtcblisting{promptbox}[1]{
  listing only,
  listing options={
    basicstyle=\ttfamily\footnotesize,
    breaklines=true,
    breakindent=0pt,
    columns=fullflexible
  },
  breakable,
  title={#1},
  fonttitle=\bfseries\small,
  coltitle=black
}

\title{As Expected? Efficient LLM Planning via Expectation--Observation Consistency Verification}

\author{
  Qingfu Zhu \quad Huanyang Zheng \quad Ting Liu \quad Wanxiang Che\thanks{Corresponding Author} \\
  {\normalfont Harbin Institute of Technology} \\
  {\normalfont\texttt{qfzhu@ir.hit.edu.cn} \quad \texttt{huanyzheng@ir.hit.edu.cn}}
}

\begin{document}
\maketitle
% Keep paragraph openings/endings together and avoid a hyphenated final line
% in the main text; restore the original settings before the references.
\begingroup
\clubpenalty=10000
\widowpenalty=10000
\finalhyphendemerits=1000000
\begin{abstract}
LLM planning often relies on reactive executors (e.g., ReAct) that generate a single action at each step to remain grounded in the latest observations, incurring substantial inference cost.
Batched executors reduce this cost by predicting multiple actions in a single model invocation, yet they are prone to compounding errors, as intermediate observations are ignored.
We propose an expectation--observation consistency verification (EOCV) mechanism for batched executors, preserving grounding while improving efficiency.
The mechanism generates expectations of action outcomes and compares them with the resulting observations.
Consistency permits continued execution, whereas mismatches trigger early termination and replanning from the current observation.
Experiments on text-based and multimodal benchmarks across multiple backbones show broadly comparable average task performance between EOCV and reactive executors, with a 29--35\% reduction in average weighted token usage.
Moreover, the mechanism is model-agnostic and remains effective when integrated into state-of-the-art planners.
\end{abstract}

\section{Introduction}
\begin{figure}[!ht]
    \centering
    \includegraphics[width=\columnwidth]{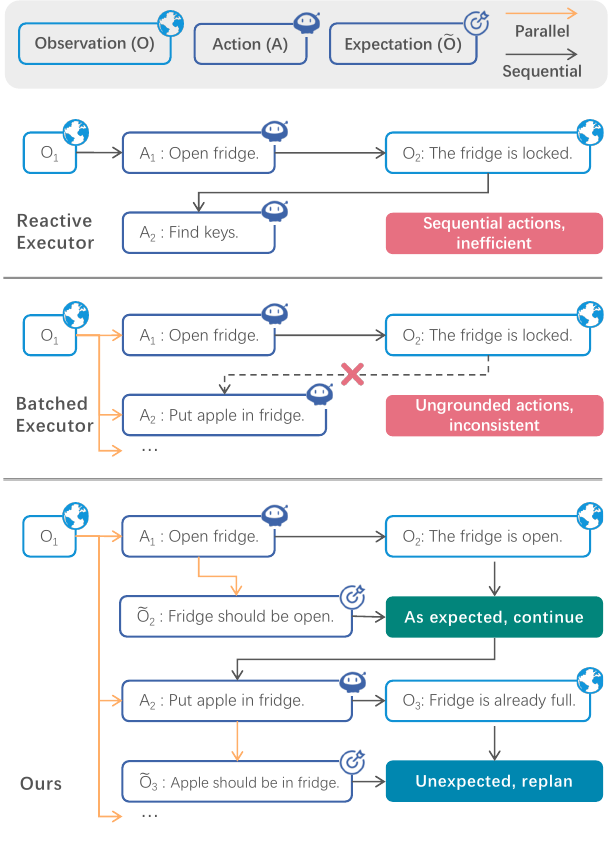}
    \caption{Executor paradigms for LLM planning.
Given the query ``Put apple in fridge'' and initial observation
$O_1$: ``The fridge is closed. You have an apple'', the reactive executor remains grounded but inefficient, whereas the batched executor is efficient but error-prone.
Ours verifies expectations against observations to continue or replan batched actions selectively.
Intermediate reasoning steps are omitted for clarity.}
    \label{fig:executor_overview}
    % Tighten only the space below Figure 1.
    \vspace{-12pt}
\end{figure}

Planning aims to generate action sequences to achieve goals under constraints and uncertainty~\cite{ghallab2004automated}.
In practice, plans need to be grounded in execution feedback~\cite{castaman2021receding,guo2023recent}, motivating an iterative observation--action loop.
ReAct~\cite{yao2022react} exemplifies this paradigm and has emerged as a foundational approach.

Existing LLM planning methods often adopt ReAct as the executor and equip it with auxiliary controller-level techniques such as task decomposition~\cite{erdoganplan}, search~\cite{hao-etal-2023-reasoning}, and reflection~\cite{shinn2023reflexion,kim-etal-2025-reflact}.
Nevertheless, the underlying ReAct execution paradigm remains unchanged: at each iteration, the agent selects a single next action given the latest observation (as shown in Figure~\ref{fig:executor_overview}).
We refer to this class of step-wise executors as \textbf{reactive executors}.
While this reactive paradigm improves robustness by remaining closely grounded in execution feedback, it causes inference costs to scale with the action sequence length, thereby limiting planning efficiency.

A natural alternative is the \textbf{batched executor}~\cite{singh2022progprompt,liu2024long}.
It generates multiple actions per model invocation, reducing repeated context prefilling and the number of internal reasoning phases.
However, batched executors weaken step-wise grounding because feedback from earlier actions is unavailable at generation time.
As a result, later actions can become inconsistent with the latest observation, increasing the risk of compounding errors.

We propose an expectation--observation consistency verification (EOCV) mechanism that combines efficient batched generation with reactive grounding.
The mechanism augments action batches with expectations of their execution outcomes.
These expectations are verified against observations as the actions are executed.
Consistency permits continued execution within the batch, whereas a mismatch triggers early termination and regeneration from the latest observation.

EOCV is model-agnostic.
It can be integrated into state-of-the-art (SOTA) planners and remains compatible with controller techniques such as task decomposition, search, and reflection.\footnote{In this paper, a planner consists of a controller and an executor; EOCV only modifies the executor and is agnostic to the controller.}
We evaluate EOCV on text-based and multimodal benchmarks, comparing against strong reactive baselines and representative batched variants.
Across multiple model backbones, EOCV reduces average weighted token usage by 29--35\%, with task performance broadly comparable to that of reactive baselines on average.
The contributions of the paper are summarized as follows:
\begin{itemize}
    \item We propose EOCV, a step-wise verification mechanism that connects batched action generation with reactive grounding.
    \item EOCV is model-agnostic and compatible with SOTA planners augmented with various controller-level techniques.
    \item Experiments on text-based and multimodal benchmarks demonstrate weighted token savings across backbones, with broadly comparable average task performance.
\end{itemize}

\begin{figure*}[ht]
    \centering
    \includegraphics[width=\textwidth]{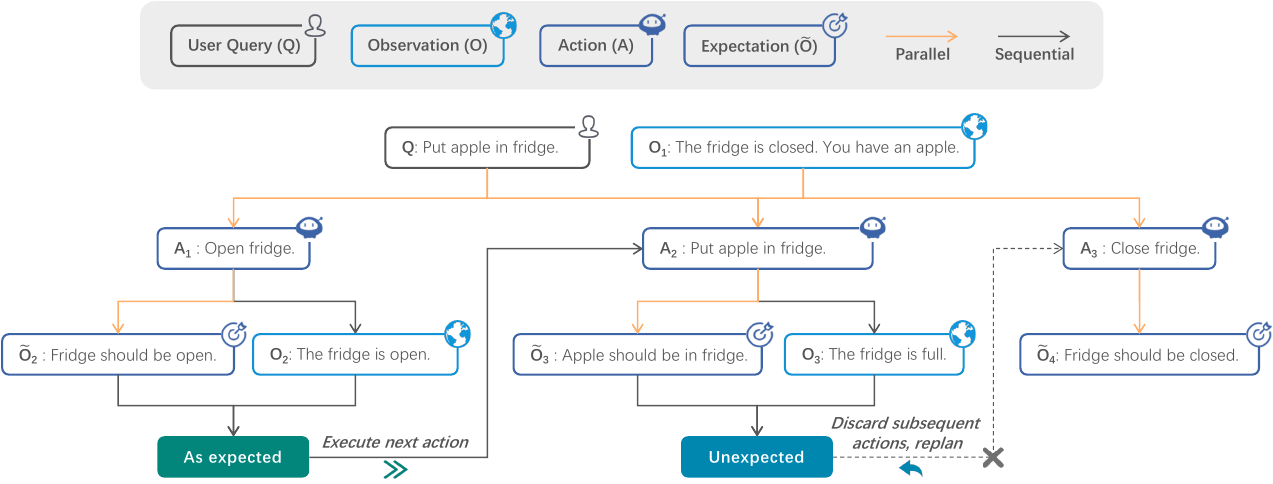}
    \caption{An overview of EOCV.
The executor generates future action batches with expectations, executes them sequentially, and verifies expectations against the resulting observations.
Consistency permits continued execution within the batch, whereas a mismatch triggers early termination and replanning from the latest observation.
}
    \label{fig:executor_method}
\end{figure*}

\section{Preliminaries}
\label{sec:preliminaries}
We study a planning executor that maps a task query $q$ to executable actions in an interactive environment.
The query can be a user instruction or a subtask produced by a planning controller.
At step $t$, the executor takes as input the current observation $o_t$ and an execution context $c_t$, and outputs action $a_t$ under policy $\pi$:
\begin{equation}
    a_t = \pi(q, o_t, c_t).
\end{equation}
The context $c_t$ summarizes the interaction history and auxiliary signals up to step $t$.
It includes past action--observation pairs and may also contain internal states such as reasoning traces.
We consider two executor paradigms: reactive executors and batched executors.

\subsection{Reactive Executor} \label{sec_react}
A reactive executor produces one action per model invocation and updates its context after each action--observation transition.
ReAct~\cite{yao2022react} is a representative instance that maintains intermediate reasoning traces within the context.
This design grounds action selection in the latest observations, but requires context prefilling and a new internal reasoning phase for each action, increasing inference cost with sequence length.

\subsection{Batched Executor}
\label{sec:batched_executor}

A batched executor generates an action sequence within a single model invocation.
At step $t$, it conditions on the query $q$, the observation $o_t$, and the context $c_t$, and generates a batch of actions autoregressively.
The context is expanded with the previously generated actions in the same batch.
\begin{equation}
\begin{aligned}
a_j &= \pi(q, o_t, c_j), \\
c_{j+1} &= c_t \oplus a_{t:j}, \qquad j = t,\ldots,t+B-1,
\end{aligned}
\end{equation}
where $a_{t:j}=(a_{t},\ldots,a_{j})$ denotes the action prefix generated so far, and $\oplus$ denotes context concatenation.
$B$ is a batch budget that upper-bounds the number of generated actions.
In practice, the realized sequence length may be determined dynamically at generation time when future environment feedback is difficult to anticipate.

Although actions are generated autoregressively, each batch uses one context prefill and one internal reasoning phase, reducing inference cost relative to reactive execution.
However, the entire batch is generated before execution begins and is then executed sequentially.
When generating $a_j$, the executor cannot condition on the intermediate observations $o_{t+1:j}$, which only become available after executing $a_{t:j-1}$.
Later actions may therefore become inconsistent with the latest observation, leading to compounding errors.

% \subsection{Batched Executor}
% \label{sec:batched_executor}

% A batched executor generates an action sequence within a single model invocation.
% At step $t$, it conditions on the current observation $o_t$ and autoregressively produces actions while expanding the context with previously generated actions:
% \begin{equation}
% \begin{aligned}
% c_j &= c_t \oplus a_{t:j-1}, \\
% a_j &= \pi(q, o_t, c_j), \qquad j = t,t+1,\ldots
% \end{aligned}
% \end{equation}
% Here $a_{t:j-1}=(a_t,\ldots,a_{j-1})$ denotes the action prefix generated so far, and $\oplus$ denotes context concatenation.
% This entire prefix is generated \emph{before} any execution begins and is then executed sequentially in the environment.
% Consequently, when generating $a_j$ for $j>t$, the executor cannot condition on the intermediate observations $o_{t+1:j}$ that would only be returned after executing $a_{t:j-1}$, which can lead to compounding errors as later actions become inconsistent with the latest observation.

\section{Method}
\label{sec:method}

To leverage the efficiency of batched executors while preserving the step-wise grounding characteristic of reactive executors,
we introduce an \textbf{expectation--observation consistency verification (EOCV)} mechanism.
As illustrated in Figure~\ref{fig:executor_method}, EOCV augments a batched executor by associating actions with explicit expectations of their execution outcomes.
These expectations are verified against the observations returned during execution.
Verification is performed step by step: success enables execution to proceed within the batch, whereas mismatches trigger early termination and regeneration from the latest observation.
We next describe expectation generation (\S\ref{sec:expected_state}) and expectation--observation verification (\S\ref{sec:verification}).

\subsection{Expectation Generation}
\label{sec:expected_state}

%To support step-wise verification, EOCV augments the executor with expectation generation.
EOCV augments the batched executor (\S\ref{sec:batched_executor}) with an explicit expectation for action $a_j$:
\begin{equation}
\hat{o}_{j+1} = \pi(q, o_t, c_j, a_j),
\end{equation}
where $\hat{o}_{j+1}$ denotes the expectation associated with action $a_j$.
We represent this expectation as a compact textual description of the action's anticipated outcome.
This expectation has the same form in both text-based and multimodal environments; verification compares it with the returned text or image observation (see \S\ref{sec:verification} for details).
Expectation generation uses only information available before execution, preserving compatibility with batched execution.

Expectation generation is unnecessary for actions whose observations are inherently unpredictable.
For information-seeking actions (e.g., \texttt{look} or \texttt{examine}), generating specific expectations increases generation and verification cost without providing a reliable consistency signal.
EOCV therefore adopts \textbf{Expectation Skipping}: these actions are executed, but their expectation generation and verification are omitted.
In the main experiments, we identify such actions with a small predefined set, which is practical for the modest action spaces in our benchmarks.
We also evaluate an automatic variant in which the executor emits an \texttt{action-is-skippable} tag to invoke skipping.
This automatic variant closely matches the predefined strategy in selecting skippable actions, downstream task performance, and efficiency.
Detailed results are reported in \S\ref{sec:ablation}.

\subsection{Expectation--Observation Verification}
\label{sec:verification}

Expectation--observation verification evaluates whether the realized environment feedback matches the expectation predicted during batched generation.
After executing action $a_j$, the environment returns an observation $o_{j+1}$, which is compared against the corresponding expectation $\hat{o}_{j+1}$ whenever an expectation is available:
\begin{equation}
v_{j+1} = V(o_{j+1}, \hat{o}_{j+1}),
\end{equation} where the function $V(\cdot)$ returns a scalar consistency score. Verification succeeds above threshold~$\tau$, whose effect is analyzed in Appendix~\ref{sec:threshold_analysis}.

If verification succeeds, execution proceeds to the next action in the batch without another executor invocation.
Otherwise, execution is terminated immediately, and the remaining actions are discarded.
This intervention limits further execution based on expectations that no longer match the environment.
EOCV benefits from batched execution by allowing multiple actions to share the same prefilled context and reasoning.
Appendix~\ref{sec:appendix_efficiency_analysis} analyzes these token~savings.

Information-seeking actions bypass verification.
Any subsequent non-information-seeking actions in the same batch are discarded, as they were generated without access to the new~information.

The implementation of $V$ depends on the observation modality.
For text observations, we use cosine similarity between expectation and observation embeddings.
For visual observations, we prompt a pretrained VLM to assess consistency.

This verification procedure assesses whether action outcomes are consistent with their expectations.
Such consistency, however, does not guarantee strategy correctness or prevent irreversible errors.
Reactive and batched baselines provide neither guarantee.

\section{Experiments}
\label{sec:experiments}
This section evaluates the effectiveness, efficiency, and generality of EOCV in interactive planning.
We first describe the benchmarks, baselines, and implementation details (\S\ref{sec:benchmarks}--\S\ref{sec:implementation}), and then present the main results and analyses (\S\ref{sec:MainResults}--\S\ref{sec:ablation}).

\subsection{Benchmarks}
\label{sec:benchmarks}

We evaluate EOCV on a benchmark suite that covers both text-based and multimodal interactive planning.
ALFWorld~\cite{ALFWorld20} and ScienceWorld~\cite{wang-etal-2022-scienceworld} provide established text-based environments with partial observability and multi-step action sequences.
EB-Habitat~\cite{yang2025embodiedbench} extends the evaluation to multimodal household operation, where the agent must act from visual observations.
The suite spans tasks from short interactions to long-horizon execution, supporting evaluation across different levels of planning complexity.

\begin{table*}[t]
\centering
\small
\setlength{\tabcolsep}{3pt}
\begin{tabular*}{\textwidth}{@{\extracolsep{\fill}}l*{7}{c}@{}}
\toprule
\multirow{2}{*}{\textbf{Method}} & \multicolumn{3}{c}{\textbf{ALFWorld}} & \multicolumn{2}{c}{\textbf{ScienceWorld}} & \multicolumn{2}{c}{\textbf{EB-Habitat}} \\
\cmidrule(lr){2-4}\cmidrule(lr){5-6}\cmidrule(lr){7-8}
 & \textbf{Reward} & \textbf{Tokens} & \textbf{Time} & \textbf{Reward} & \textbf{Tokens} & \textbf{Reward} & \textbf{Tokens} \\
\midrule
Reactive & 88.81 & 1.00$\times$ & 1.00$\times$ & 89.37 & 1.00$\times$ & 66.94 & 1.00$\times$ \\
Batched & 83.58 & 0.57$\times$ & 0.53$\times$ & 82.56 & 0.41$\times$ & 56.56 & 0.38$\times$ \\
Ours & 88.56 & 0.65$\times$ & 0.59$\times$ & 88.87 & 0.69$\times$ & 67.77 & 0.71$\times$ \\
\bottomrule
\end{tabular*}
\caption{Average executor results on ALFWorld, ScienceWorld, and EB-Habitat.
Reward is the mean score on $[0,100]$; Tokens denotes weighted token usage; Time is summed per-task wall-clock time.
Token and time ratios are normalized to Reactive ($1.00\times$) within each backbone before averaging. ALFWorld time averages cover the locally deployed Qwen3.5-4B and Qwen3.5-9B models.
\label{tab:main_results}}
\end{table*}

\subsection{Baselines}
\label{sec:baselines}

Following the executor--controller formulation in \S\ref{sec:preliminaries}, we evaluate baselines at two levels.
At the executor level, we evaluate two primary baselines: a ReAct-based reactive executor~\cite{yao2022react} and a batched executor.
The batched executor generates multiple actions per model invocation.
Both baselines share the executor implementation used by EOCV, enabling a direct comparison of execution strategies.
We additionally evaluate AdaPlanner~\cite{sun2023adaplanner}, which uses code assertions to check subgoal completion and triggers plan refinement when these checks fail.
Its implementation and shared in-context example are detailed in Appendices~\ref{sec:appendix_adaplanner} and~\ref{sec:prompt_adaplanner}.
We also evaluate a batched variant that replans when the environment reports an action failure, without generating expectations (Fail-RP).
At the planner level, we use SOTA planner baselines that pair reactive executors with three representative controller techniques: ReflAct, Reflexion, and MPO.
These baselines allow us to evaluate whether EOCV remains effective in stronger planning pipelines.

\paragraph{ReflAct.}
This controller technique provides an inline enhancement for reactive executors by shifting executor reasoning from merely planning the next action to continuously reflecting on the agent's state relative to its goal~\cite{kim-etal-2025-reflact}.

\paragraph{Reflexion.}
This controller technique augments reactive executors with a memory mechanism that stores feedback from previous episodes to refine subsequent executions~\cite{shinn2023reflexion}.

\paragraph{MPO.}
This controller technique decomposes high-level tasks into structured subgoals and provides step-level guidance to the executor~\cite{xiong-etal-2025-mpo}.

\subsection{Implementation Details}
\label{sec:implementation}

The main executor comparisons use Qwen3.5-4B, Qwen3.5-9B, and DeepSeek-V3.2~\cite{liu2025deepseek} on ALFWorld; GPT-5.5, DeepSeek-V4.1-Flash, and DeepSeek-V3.2 on ScienceWorld; and Qwen3.6-Flash and Qwen3.6-Plus on EB-Habitat.
We use greedy decoding with reasoning disabled for the DeepSeek models, Qwen3.6-Flash, and Qwen3.6-Plus.
For GPT-5.5, we set the reasoning effort to \texttt{low}.
For Qwen3.5-4B and Qwen3.5-9B, we set enable-thinking to \texttt{on}.
For verification (\S\ref{sec:verification}), text-based experiments use the all-mpnet-base-v2 sentence-transformer model~\cite{reimers-2019-sentence-bert} with a cosine threshold of 0.6, and multimodal experiments use the executor backbone as the prompted VLM verifier.
Batched and EOCV-enhanced executors generate at most five future actions per model invocation.
Reflexion uses five iterations on ALFWorld and three each on ScienceWorld \mbox{and EB-Habitat}.

We report wall-clock time only for Qwen3.5-4B and Qwen3.5-9B, each deployed locally using vLLM on a single NVIDIA A100-SXM4-80GB GPU.
Local deployment enables more controlled timing comparisons by avoiding network and API server variability.

We follow the standard evaluation protocol for each benchmark.
For ALFWorld, we evaluate the full unseen set of 134 tasks with a budget of 50 actions per task.
For ScienceWorld, we follow~\citet{lin2023swiftsage} and evaluate the first 10 test instances of each task type, or all available instances when fewer than 10 are provided, yielding 271 instances with a budget of 70 actions per instance.
For EB-Habitat, we evaluate 50 tasks in each of the six categories, for a total of 300 tasks.
For MPO, we restrict evaluation to instances with meta-plans available in the official repository: all ALFWorld tasks and 211 of the 271 ScienceWorld instances.
We do not evaluate MPO on EB-Habitat because no meta-plans are available.
Appendix~\ref{sec:appendix_prompt} provides the prompts and in-context examples for these~benchmarks.

\subsection{Evaluation Metrics}
\label{sec:metrics}

% Queue the wide table early so it shares a page with the verifier analysis.
\begin{table*}[t]
\centering
\scriptsize
\resizebox{\textwidth}{!}{
\begin{tabular}{lcccc|cccc|cc}
\toprule
\multirow{2}{*}{\textbf{Verifier}} &
\multicolumn{4}{c|}{\textbf{ALFWorld}} &
\multicolumn{4}{c|}{\textbf{ScienceWorld}} &
\multicolumn{2}{c}{\textbf{EB-Habitat}} \\
\cmidrule(lr){2-5}\cmidrule(lr){6-9}\cmidrule(lr){10-11}
& \textbf{Prec.} & \textbf{Rec.} & \textbf{Reward} & \textbf{Tokens}
& \textbf{Prec.} & \textbf{Rec.} & \textbf{Reward} & \textbf{Tokens}
& \textbf{Prec.} & \textbf{Rec.} \\
\midrule
Embedding
& 100.00 & 100.00 & 94.78 & 1.00$\times$
& 96.46 & 100.00 & 85.42 & 1.00$\times$
& -- & -- \\
LLM/VLM
& 100.00 & 100.00 & 94.78 & 1.06$\times$
& 97.32 & 100.00 & 85.93 & 1.15$\times$
& 89.38 & 87.07 \\
\bottomrule
\end{tabular}}
\caption{Verifier comparison. Precision (Prec.) and recall (Rec.) use 300 randomly sampled,
manually labeled pairs per benchmark. Pairs are collected with DeepSeek-V3.2 on the text benchmarks,
while Qwen3.6-Flash is used for EB-Habitat. Tokens are normalized to embedding verification within each text benchmark;
dashes mark inapplicable entries.\label{tab:verifier_analysis}}
\end{table*}

Task performance is reported as \textbf{reward} on a unified $[0,100]$ scale.
The underlying environment metrics are binary success for ALFWorld, graded progress for ScienceWorld, and task-progress score for EB-Habitat, with the latter rescaled to $[0,100]$.
We normalize scores where necessary and report averages over evaluation episodes.
Execution efficiency is measured by a \textbf{weighted token count}.
We aggregate prompt and completion tokens across all evaluation episodes and weight them by their pricing.
Detailed weights are listed in Appendix~\ref{sec:appendix_token_accounting}.

\begin{figure}[t]
    \centering
    \includegraphics[width=\columnwidth]{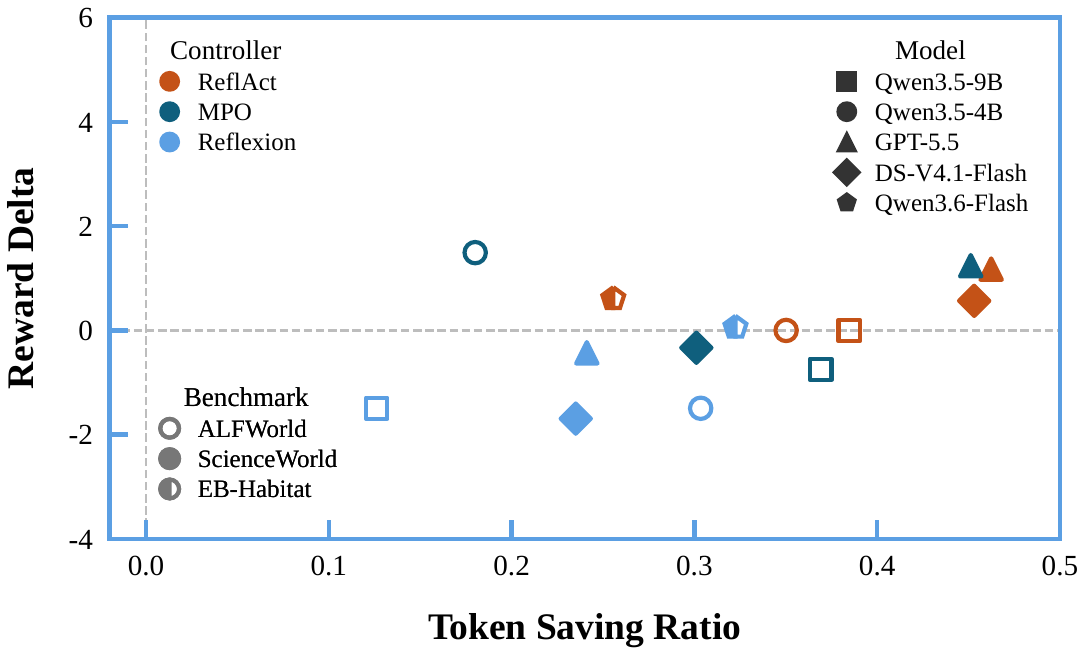}
    \caption{EOCV with different planner controllers. The x-axis shows the proportional reduction in weighted token usage, and the y-axis shows the change in reward, both relative to the corresponding planner with a reactive executor. Colors indicate controllers, and marker shapes indicate backbone models. Hollow, filled, and half-filled markers indicate ALFWorld, ScienceWorld, and EB-Habitat, respectively.}
    \label{fig:agnostic}
\end{figure}

\subsection{Main Results}
\label{sec:MainResults}
Table~\ref{tab:main_results} compares executor variants without controllers.
Across ALFWorld, ScienceWorld, and EB-Habitat, the reactive executor remains well grounded but uses the most tokens, whereas the batched executor lowers cost with a clear performance penalty.
EOCV achieves broadly comparable average task performance to the reactive executor while reducing average weighted token usage to 0.65$\times$, 0.69$\times$, and 0.71$\times$ on the three benchmarks, respectively.
ALFWorld wall-clock time falls to 0.59$\times$ on average across the two locally deployed Qwen models.

Figure~\ref{fig:agnostic} evaluates EOCV in combination with planner controllers across text-based and multimodal benchmarks.
Across the evaluated controllers, EOCV consistently reduces weighted token usage while maintaining broadly comparable task performance overall.
These results suggest that EOCV can serve as an executor component across planners without modifying their controllers.
Table~\ref{tab:controller_composition} in Appendix~\ref{sec:appendix_additional_results} reports detailed reward and token usage for the controller comparisons across all three benchmarks.

Across 28 equally weighted paired configurations, the mean reward difference between EOCV and the corresponding reactive baselines is $+0.0458$ points (95\% confidence interval based on the paired $t$ procedure: [$-0.3406$, $+0.4321$]).
A two-sided paired $t$-test finds no statistically significant difference in mean reward ($p=0.8098$).
The near-zero observed mean difference is consistent with broadly comparable average task performance across the evaluated configurations.
Details are provided in Appendix~\ref{sec:appendix_reward_summary}.

\subsection{Analysis}
\label{sec:ablation}
The analysis first assesses the effectiveness and reliability of EOCV's joint expectation--verification mechanism.
It then examines the efficiency gains from expectation skipping and component contributions through progressive ablation.
Finally, case studies illustrate execution behavior, and sensitivity analysis evaluates parameter effects.

\begin{figure}[t]
    \centering
    % Generated by scripts/plot_additional_baselines.py --layout panels --compact.
    \includegraphics[width=\columnwidth]{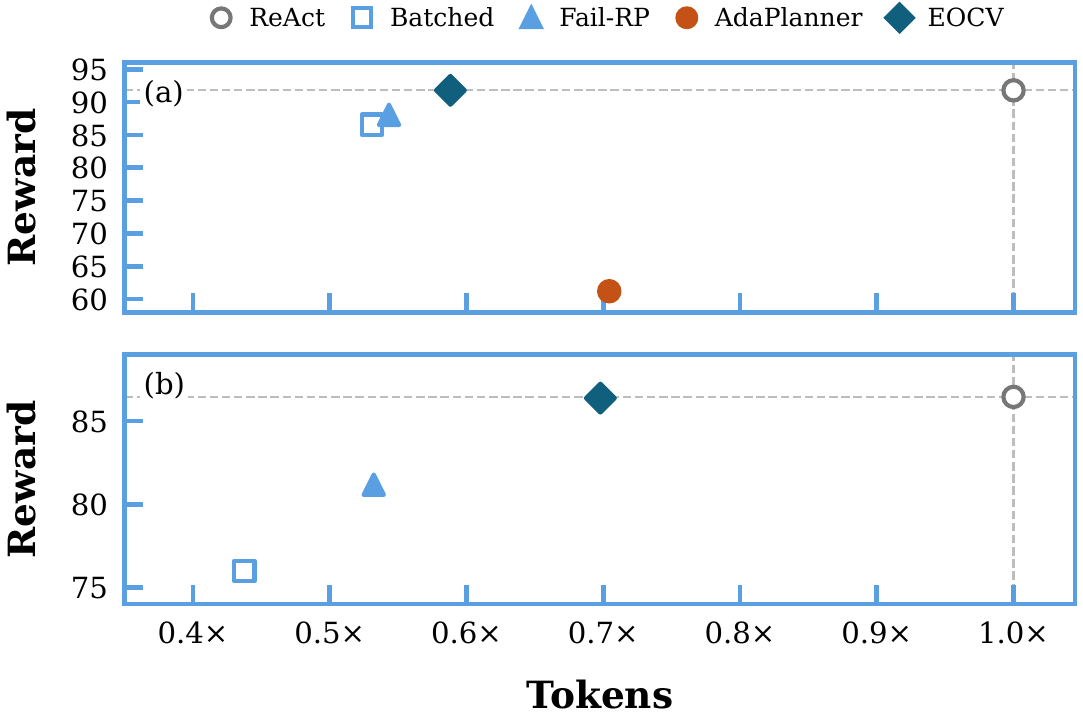}
    \caption{Executor comparisons on (a) ALFWorld with Qwen3.5-9B and (b) ScienceWorld with DeepSeek-V4.1-Flash. Reward is the mean score on $[0,100]$; Tokens denotes weighted token usage normalized to ReAct ($1.00\times$) within each setting. AdaPlanner is evaluated only on ALFWorld. Dashed lines mark the corresponding ReAct baseline.}
    \label{fig:additional_baselines}
\end{figure}

\paragraph{Effectiveness and Efficiency.}
Figure~\ref{fig:additional_baselines} evaluates explicit expectation--verification across representations and granularities in text-based environments.
Fail-RP relies solely on environment failure signals, whereas AdaPlanner encodes expectations and verification as executable assertions for subgoals that may span multiple actions.
EOCV combines natural-language expectations with semantic verification at the action level.
It uses more weighted tokens than Fail-RP but increases reward by 3.73 and 5.21 points on ALFWorld and ScienceWorld, respectively, reaching performance comparable to ReAct.
This performance is achieved with 41\% and 30\% fewer weighted tokens than ReAct, respectively, offering a favorable balance between reward and token usage.
On ALFWorld, EOCV also achieves higher reward with fewer tokens than AdaPlanner.
These baselines depend on explicit failure feedback or states accessible to code assertions; EOCV supports verification from visual observations when such information is unavailable.

% Queue the analysis tables early so they can share pages with their discussion.
\begin{table*}[t]
\centering
\scriptsize
\setlength{\tabcolsep}{3pt}
\renewcommand{\arraystretch}{1.12}

\begin{tabularx}{\textwidth}{@{}c Y Y Y Y Y@{}}
\toprule
 & \multicolumn{2}{c}{\textbf{Batched Executor}}
 & \multicolumn{3}{c}{\textbf{Ours}} \\
\cmidrule(lr){2-3} \cmidrule(lr){4-6}

\textbf{Step} & \textbf{Action} & \textbf{Observation}
              & \textbf{Action} & \textbf{Expectation} & \textbf{Observation} \\
\midrule

$t$ &
\texttt{move(glass cup, oven)} &
That can't be moved there, because the oven isn't open. &
\texttt{move(glass cup, oven)} &
You move the glass cup to the oven. &
That can't be moved there, because the oven isn't open. \\

$t{+}1$ &
\textcolor{casebaseline}{\texttt{activate(oven)}} &
The oven is now activated. &
\textcolor{casewithheld}{\texttt{close(oven)}} &
The oven is now closed.  &
\textcolor{caseeocv}{\textbf{Terminate \& Replan}} \\

\bottomrule
\end{tabularx}

\caption{ScienceWorld case study with DeepSeek-V4.1-Flash. Steps are aligned at each executor's failed move.
\textcolor{casebaseline}{Red} denotes batched actions executed after the failure; \textcolor{casewithheld}{gray} denotes actions withheld by EOCV.
The full batches are shown in Table~\ref{tab:appendix_text_case}.}

\label{tab:case_study_executor}
\end{table*}

\begin{table*}[t]
\centering
\scriptsize
\setlength{\tabcolsep}{3pt}
\renewcommand{\arraystretch}{1.05}

\begin{tabularx}{\textwidth}{@{}c Y I A Y E I A@{}}
\toprule
 & \multicolumn{3}{c}{\textbf{Batched Executor}}
 & \multicolumn{4}{c}{\textbf{Ours}} \\
\cmidrule(lr){2-4} \cmidrule(lr){5-8}

\textbf{Step} & \textbf{Action} & \textbf{Observation} & \textbf{Annotation}
              & \textbf{Action} & \textbf{Expectation} & \textbf{Observation} & \textbf{Annotation} \\

\midrule

$t$ &
\texttt{pick up the pear} &
\includegraphics[width=0.07\textwidth]{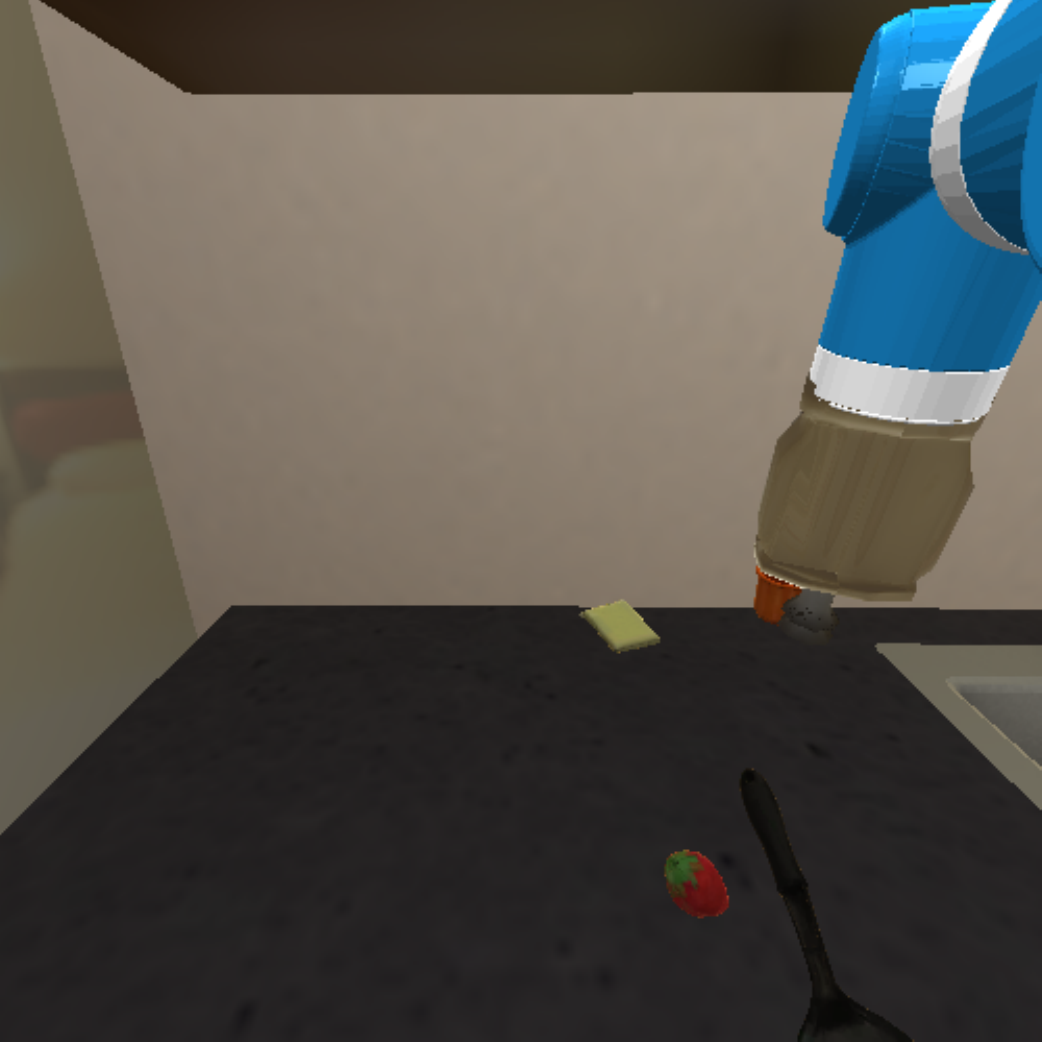} &
nothing is picked up &
\texttt{pick up the pear} &
The robot picks up a pear from the left counter. &
\includegraphics[width=0.07\textwidth]{figures/eb-habitat-counter.pdf} &
nothing is picked up \\

$t{+}1$ &
\textcolor{casebaseline}{\texttt{navigate to the sofa}} &
\includegraphics[width=0.07\textwidth]{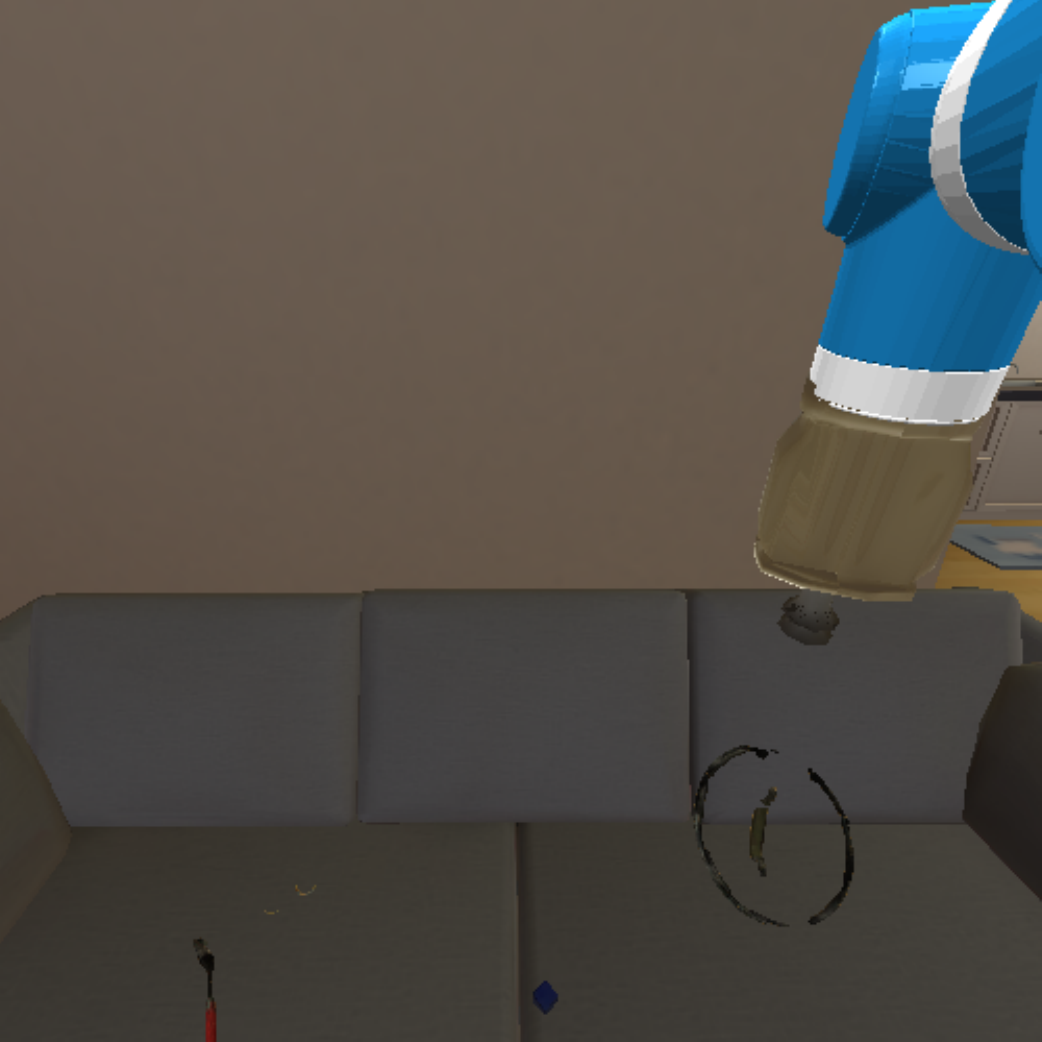} &
agent moves to the sofa &
\textcolor{casewithheld}{\texttt{navigate to the sofa}} &
\textcolor{caseeocv}{\textit{Skip verification}} &
\textcolor{caseeocv}{\textbf{Terminate \& Replan}} &
--- \\

\bottomrule
\end{tabularx}

\caption{Multimodal EB-Habitat case study with Qwen3.6-Flash. The batched executor continues with a stale plan after the shared pick-up step, whereas EOCV detects an expectation--observation mismatch and triggers replanning. Annotations describe visual observations for human readers and are not accessible to the agent. The full trajectory is reported in Table~\ref{tab:appendix_multimodal_case}; dashes indicate not applicable.}
\label{tab:case_study_multimodal}
\end{table*}

\begin{table}[t]
\centering
\small
\begin{tabular}{lcc}
\toprule
Setting
& Reward $\uparrow$
& Tokens (M) $\downarrow$ \\
\midrule
\textbf{Full Model}
& 94.78& 2.45\\

w/o Expectation Skipping
& 94.78& 3.96\\

% w/o Dynamic Batch Length
% & 94.03& 4.16\\

w/o Verification
& 90.30& 1.73\\

%w/o Batched (\textit{ReAct})
%& 94.78& 4.02\\
\bottomrule
\end{tabular}

% \vspace{0.2em}
% \raggedright
% \footnotesize
% Components are removed cumulatively; w/o Verification also removes expectation generation.
% Token usage is weighted by execution length.
\caption{Progressive ablation on ALFWorld with DeepSeek-V3.2. Components are removed cumulatively; w/o Verification omits both expectation generation and verification.\label{tab:ablation}}
\end{table}

\paragraph{Verifier Reliability.}
Verifier judgments determine whether EOCV continues a batch or replans.
Averaged across backbone--benchmark settings, observations are judged consistent with their corresponding expectations for 66.78\% of non-skipped classified actions in text-based environments and 28.55\% in multimodal environments.
Observations judged inconsistent trigger early termination and replanning (Appendix~\ref{sec:appendix_verification_outcomes}).

To assess verifier reliability, two graduate students independently label 300 randomly sampled expectation--observation pairs per benchmark and resolve initial disagreements (approximately 0.67\% of all pairs) through discussion.
On text benchmarks, embedding verification achieves precision, recall, and task reward comparable to the Qwen3.5-Flash LLM verifier with lower weighted token usage (Table~\ref{tab:verifier_analysis}).
These results favor embedding verification as an efficient alternative to LLM verification for text observations and support VLM verification for visual observations.
Appendix~\ref{sec:appendix_verifier_difficulty} provides further analysis of verification difficulty on EB-Habitat.

Occasional verifier errors do not necessarily lead to task failure.
Rejecting a consistent outcome triggers additional replanning, whereas accepting an inconsistent outcome can delay it.
Subsequent verification may still detect deviations and prompt replanning from the latest observation.
This loop provides opportunities to limit the effects of individual verification errors on subsequent execution.

\paragraph{Skipping Strategy.}
Expectation skipping avoids unnecessary expectation generation and verification for information-seeking actions.
The main experiments use predefined rules because these actions are readily identifiable within the benchmarks' small action spaces.
Skipping rates average 64.0\% in text-based environments and 59.0\% in multimodal environments across backbone--benchmark settings.
Appendix~\ref{sec:appendix_verification_outcomes} provides detailed distributions.

The executor can also predict these skipping decisions to reduce manual specification.
On 400 randomly sampled actions from DeepSeek-V3.2 runs on ALFWorld, automatic skipping matches rule-based decisions with precision and recall above 99\%.
At the task level, the same strategy yields a similar reward of 94.03 versus 94.78 while using 0.92 times as many weighted tokens.
These results support automatic skipping as an alternative in this setting.
Appendix~\ref{sec:appendix_skip_verify} provides detailed results and a skip-only ablation.

\paragraph{Ablation Study.}
Table~\ref{tab:ablation} reports a progressive ablation following component dependencies.
Disabling expectation skipping increases weighted token usage while reward remains unchanged, indicating that skipping improves efficiency.
Expectation generation and verification are then ablated as a unified mechanism because the generated expectations provide the reference for consistency verification.
Token usage decreases but reward drops, indicating that this mechanism helps maintain task performance during batched execution.

% Further removing dynamic batch length (\S\ref{sec:batched_executor}) leads to a further increase in token usage.
% With a fixed batch length, the executor cannot adapt how many actions to generate as the environment evolves, resulting in unnecessary generation once future outcomes become uncertain, while reward remains comparable.

\paragraph{Case Study.}
Tables~\ref{tab:case_study_executor} and~\ref{tab:case_study_multimodal} show representative text-based and multimodal failure modes of batched executors.
After \texttt{move(glass cup, oven)} fails because the oven is closed, the batched executor continues with its planned activation, waiting, and inspection actions.
EOCV discards the remaining wait and inspection actions and replans from the latest observation.
% 这个“类似的故障模式”描述还要不要更详细点？
A similar failure pattern appears in the EB-Habitat example with visual observations.
The executor initially attempts to \texttt{pick up the pear}, but the following image observation contradicts the corresponding expectation, since the pear is not present in the scene.
The EOCV-enhanced executor detects this mismatch, discards the remaining actions, and replans from the latest observation.
Complete versions of both examples are provided in Appendix~\ref{sec:appendix_case_studies}.

\paragraph{Hyperparameter Sensitivity.}
Average reward remains largely stable across the evaluated verification thresholds and batch budgets.
Larger budgets reduce token usage with diminishing gains, while threshold choice has a modest effect on token usage (Appendix~\ref{sec:appendix_hyperparameters}).

\section{Related Work}
% We review planning methods at the executor and controller levels.

% Planning Executors represents the "llm as planner" paradigm in the field of LLM planning.
% These executors generates actions independently, while other agents translate natual language to formal planning languages(e.g. PDDL), and use a classical planner to solve the planning task.
% The latter is known as the "llm as translator" paradigm.

\subsection{Reactive Executor}

% Reactive executors are designed to address both traditional question-answering tasks, such as HotpotQA~\cite{yang-etal-2018-hotpotqa} and GSM8K~\cite{cobbe2021gsm8k}, as well as multi-step interactive environments, including ALFWorld.
% Reactive executors operate in an iterative manner, interleaving action generation with context accumulation to adaptively plan subsequent steps.

Reactive executors follow a closed-loop \emph{reason--act--feedback} paradigm.
SayCan~\cite{ahn2022can} samples candidate robot actions from an LLM and selects one using a value function.
Inner Monologue~\cite{huang2022inner} incorporates environment descriptions into its reasoning as action feedback.
ReAct~\cite{yao2022react} makes reasoning explicit in this execution loop.

Selecting each next action requires another model invocation, repeating context prefilling and internal reasoning and increasing inference cost.

\subsection{Batched Executor}

% Batched executors represent an alternative execution paradigm, in which a future action sequence is generated within a single LLM invocation.
% By eliminating the need to repeatedly recompute and condition on the execution context after each action, batched generation substantially improves execution efficiency; however, it also demands a stronger model-level understanding of the environment, as the executor must anticipate future state transitions without intermediate feedback.

Batched executors were initially explored in question-answering settings and later extended to interactive and embodied tasks.
ReWOO~\cite{ReWoo2023} separates reasoning from tool observations, while Plan-and-Solve Prompting~\cite{wang-etal-2023-plan} decomposes tasks before sequential execution.
ProgPrompt~\cite{singh2022progprompt} uses conditional programs for embodied tasks.

To retain grounding during multi-step execution, AdaPlanner~\cite{sun2023adaplanner} uses code assertions to verify subgoals that may contain multiple actions, drawing on textual feedback and HTML states.
HVR~\cite{petruzzellis2025hierarchical} validates plans and monitors execution using symbolic states and PDDL action conditions.
More recently, SPACE~\cite{yang2026space} learns to adapt action-chunk length using boundary supervision from programmatic skills, while CheckVLA~\cite{liu2026checkvla} verifies vision--language--action execution with a trained action-conditioned world model.
% Like receding horizon planning~\cite{castaman2021receding}, EOCV revises plans using execution feedback.
% It couples expectation generation with semantic consistency verification, comparing expectations against textual or visual observations to continue the batch or replan without a manually specified symbolic transition model.
EOCV shares the principle of revising plans from execution feedback with receding horizon planning~\cite{castaman2021receding}.
It extends this principle to batched LLM agents by coupling expectation generation with semantic consistency verification.
Explicit expectations provide the reference for verification.
The verifier compares these expectations with textual or visual observations to determine whether to continue the batch or replan.
This verification does not require a manually specified symbolic transition model.

% Batched Executor stands for another different approach, which yields a future action sequence with a single LLM invocation.
% Batched generation makes these executors more efficient, but also requires a deeper understanding of the environment, as they have to predict the possible environment state.
% Similar to reactive executors, batched executors are first developed in QA tasks, then applied to interactive or embodied tasks.
% ReWOO(Reasoning Without Observation)\cite{ReWoo2023} detaches the reasoning process from external observations using a blueprint of an interdependent plan, thus significantly reducing token consumption.
% Plan-And-Solve Prompting\cite{wang-etal-2023-plan} devising a plan to divide the entire task into smaller subtasks, and then carrying out the subtasks according to the plan. This outperforms Zero-shot-CoT on math-related QA tasks.
% ProgPrompt\cite{singh2022progprompt} introduces Batched Executor into the embodied intelligence field by presenting a programmatic LLM prompt structure. This structure is equipped with conditional control and can be executed to complete the task.
% Although these executors work well on former benchmarks, they struggle on benchmarks with complex interactions and flexible environments, such as ALFWorld and Science World.

\subsection{Planning Controller}
Controllers complement executors through task decomposition, search, and reflection.
% Task decomposition is one of the most widely adopted controller paradigms, in which a complex task is decomposed into a set of manageable subtasks.
Multi-agent approaches such as Reason--Plan--ReAct~\cite{molinari2025reason} introduce a high-level controller responsible for subtask segmentation.
Graph-based methods including HyperTree Planning~\citep{gui2025hypertree} leverage structured data representations to support dynamic task expansion and iterative refinement.
% Another line of work focuses on context management, where additional documents, experiences, or memories are supplied to augment the executor.
Context management methods such as AutoManuals~\citep{cai2025beyond} provide the executor with detailed environment exploration traces.
% whereas Dynamic Cheatsheet~\citep{suzgun2025dynamic} maintains cross-task memories and reusable strategies.
% Self-reflection has also emerged as a promising controller mechanism, enabling agents to improve execution by learning from prior failures.
The self-improvement framework Meta-Policy Reflexion~\citep{wu2025meta} stores and applies reflective feedback through a dual-track policy architecture.
% Meta-Policy Reflexion~\citep{wu2025meta} stores reflective feedback and applies it through a dual-track policy architecture, while STeP~\citep{chen2025training} synthesizes self-reflective trajectories to more effectively distill supervision from teacher models.

Despite their architectural diversity, most existing controllers rely on similar reactive executors, allowing EOCV to replace the executor without modifying the controller.
% Our approach replaces the core executor while keeping compositionality with the controllers, achieving substantial token efficiency improvements without sacrificing overall performance.

\section{Conclusion}
\label{sec:conclusion}

This work introduces EOCV to combine efficient batched generation with reactive grounding through expectation--observation consistency verification.
Verification determines whether to continue the batch or replan from the latest observation.
Across ALFWorld, ScienceWorld, and EB-Habitat, EOCV reduces average weighted token usage by 29--35\% while maintaining average task performance comparable to reactive execution.
Token savings persist with different planner controllers across language and vision-language backbones, supporting EOCV as a model-agnostic approach to efficient interactive planning.

% Controllers are enhancement modules working with executors. They originate from different fields and optimize the executor from different perspectives.
% Task decomposition is a common controller method which splits main task into subtasks.
% Multi-agent methods like Reason-Plan-ReAct\cite{molinari2025reason} train a high-level Planner to separate subtasks while keeping ReAct as the low-level executor.
% Other methods such as HyperTree Planning~\citep{gui2025hypertree} and ReAcTree~\citep{choi2025reactree} deployed data structures to manage dynamic task expansion and refinement.
% Context Management methods provide extra documents, experiences or memories to enhance executors.
% AutoManuals~\citep{cai2025beyond} gives out a detailed environment exploration record, while Dynamic Cheatsheet~\citep{suzgun2025dynamic} offers cross-task memories and strategies.
% Self Reflection is also a promising method, which learns from previous failure to improve execution.
% Meta-Policy Reflexion\cite{wu2025meta} store agent reflection and apply them using a dual-track system.
% STeP\cite{chen2025training} synthesizes self-reflected trajectories to enhance the effectiveness in learning from teacher models.
% Although these controllers vary in form, their executors are mostly the same. Our method aims at replacing the executor core while keeping the controller unchanged, and provides considerable token efficiency while maintaining performance.

\newpage
\subsection*{Limitations}
\label{sec:limitations}

Verification assesses consistency between observed action outcomes and their corresponding expectations.
Consistency alone does not establish the suitability of the underlying strategy for the task.
Verification occurs after action execution.
Replanning can therefore revise subsequent actions but cannot reverse the effects of irreversible actions.
Addressing these limitations would require additional mechanisms for assessing progress toward the task goal and evaluating action consequences before execution.

\par\endgroup

\bibliography{custom}

\begin{thebibliography}{29}
\providecommand{\natexlab}[1]{#1}

\bibitem[{Ahn et~al.(2022)Ahn, Brohan, Brown, Chebotar, Cortes, David, Finn,
  Fu, Gopalakrishnan, Hausman et~al.}]{ahn2022can}
Michael Ahn, Anthony Brohan, Noah Brown, Yevgen Chebotar, Omar Cortes, Byron
  David, Chelsea Finn, Chuyuan Fu, Keerthana Gopalakrishnan, Karol Hausman, and
  1 others. 2022.
\newblock Do as i can, not as i say: Grounding language in robotic affordances.
\newblock \emph{arXiv preprint arXiv:2204.01691}.

\bibitem[{Cai et~al.(2025)Cai, Liu, Yang, Niu, Xiao, and Chen}]{cai2025beyond}
Kuntai Cai, Juncheng Liu, Xianglin Yang, Zhaojie Niu, Xiaokui Xiao, and Xing
  Chen. 2025.
\newblock Beyond manuals and tasks: Instance-level context learning for llm
  agents.
\newblock \emph{arXiv preprint arXiv:2510.02369}.

\bibitem[{Castaman et~al.(2021)Castaman, Pagello, Menegatti, and
  Pretto}]{castaman2021receding}
Nicola Castaman, Enrico Pagello, Emanuele Menegatti, and Alberto Pretto. 2021.
\newblock Receding horizon task and motion planning in changing environments.
\newblock \emph{Robotics and Autonomous Systems}, 145:103863.

\bibitem[{Erdogan et~al.(2025)Erdogan, Lee, Kim, Moon, Furuta, Anumanchipalli,
  Keutzer, and Gholami}]{erdoganplan}
Lutfi~Eren Erdogan, Nicholas Lee, Sehoon Kim, Suhong Moon, Hiroki Furuta,
  Gopala Anumanchipalli, Kurt Keutzer, and Amir Gholami. 2025.
\newblock \href {https://proceedings.mlr.press/v267/erdogan25a.html}
  {Plan-and-act: Improving planning of agents for long-horizon tasks}.
\newblock In \emph{Proceedings of the 42nd International Conference on Machine
  Learning}, volume 267 of \emph{Proceedings of Machine Learning Research},
  pages 15419--15462. PMLR.

\bibitem[{Ghallab et~al.(2004)Ghallab, Nau, and
  Traverso}]{ghallab2004automated}
Malik Ghallab, Dana Nau, and Paolo Traverso. 2004.
\newblock \emph{Automated Planning: theory and practice}.
\newblock Elsevier.

\bibitem[{Gui et~al.(2025)Gui, Wang, Wang, Ma, Zhen, Yuan, Hao, Lian, Chen, and
  Wu}]{gui2025hypertree}
Runquan Gui, Zhihai Wang, Jie Wang, Chi Ma, Huiling Zhen, Mingxuan Yuan, Jianye
  Hao, Defu Lian, Enhong Chen, and Feng Wu. 2025.
\newblock \href {https://proceedings.mlr.press/v267/gui25b.html} {{H}yper{T}ree
  planning: Enhancing {LLM} reasoning via hierarchical thinking}.
\newblock In \emph{Proceedings of the 42nd International Conference on Machine
  Learning}, volume 267 of \emph{Proceedings of Machine Learning Research},
  pages 20694--20722. PMLR.

\bibitem[{Guo et~al.(2023)Guo, Wu, Qin, Li, Li, and Li}]{guo2023recent}
Huihui Guo, Fan Wu, Yunchuan Qin, Ruihui Li, Keqin Li, and Kenli Li. 2023.
\newblock Recent trends in task and motion planning for robotics: A survey.
\newblock \emph{ACM Computing Surveys}, 55(13s):1--36.

\bibitem[{Hao et~al.(2023)Hao, Gu, Ma, Hong, Wang, Wang, and
  Hu}]{hao-etal-2023-reasoning}
Shibo Hao, Yi~Gu, Haodi Ma, Joshua Hong, Zhen Wang, Daisy Wang, and Zhiting Hu.
  2023.
\newblock \href {https://doi.org/10.18653/v1/2023.emnlp-main.507} {Reasoning
  with language model is planning with world model}.
\newblock In \emph{Proceedings of the 2023 Conference on Empirical Methods in
  Natural Language Processing}, pages 8154--8173, Singapore. Association for
  Computational Linguistics.

\bibitem[{Huang et~al.(2023)Huang, Xia, Xiao, Chan, Liang, Florence, Zeng,
  Tompson, Mordatch, Chebotar, Sermanet, Brown, Jackson, Luu, Levine, Hausman,
  and Ichter}]{huang2022inner}
Wenlong Huang, Fei Xia, Ted Xiao, Harris Chan, Jacky Liang, Pete Florence, Andy
  Zeng, Jonathan Tompson, Igor Mordatch, Yevgen Chebotar, Pierre Sermanet, Noah
  Brown, Tomas Jackson, Linda Luu, Sergey Levine, Karol Hausman, and Brian
  Ichter. 2023.
\newblock Inner monologue: Embodied reasoning through planning with language
  models.
\newblock In \emph{Proceedings of The 6th Conference on Robot Learning}.

\bibitem[{Kim et~al.(2025)Kim, Rhee, Kim, Kim, Lee, Sung, and
  Jung}]{kim-etal-2025-reflact}
Jeonghye Kim, Sojeong Rhee, Minbeom Kim, Dohyung Kim, Sangmook Lee, Youngchul
  Sung, and Kyomin Jung. 2025.
\newblock \href {https://doi.org/10.18653/v1/2025.emnlp-main.1697}
  {{R}efl{A}ct: World-grounded decision making in {LLM} agents via goal-state
  reflection}.
\newblock In \emph{Proceedings of the 2025 Conference on Empirical Methods in
  Natural Language Processing}, pages 33421--33453, Suzhou, China. Association
  for Computational Linguistics.

\bibitem[{Lin et~al.(2023)Lin, Fu, Yang, Brahman, Huang, Bhagavatula,
  Ammanabrolu, Choi, and Ren}]{lin2023swiftsage}
Bill~Yuchen Lin, Yicheng Fu, Karina Yang, Faeze Brahman, Shiyu Huang, Chandra
  Bhagavatula, Prithviraj Ammanabrolu, Yejin Choi, and Xiang Ren. 2023.
\newblock \href {https://doi.org/10.52202/075280-1034} {Swiftsage: A generative
  agent with fast and slow thinking for complex interactive tasks}.
\newblock In \emph{Advances in Neural Information Processing Systems},
  volume~36, pages 23813--23825. Curran Associates, Inc.

\bibitem[{Liu et~al.(2025)Liu, Mei, Lin, Xue, Wang, Xu, Wu, Zhang, Lin, Dong
  et~al.}]{liu2025deepseek}
Aixin Liu, Aoxue Mei, Bangcai Lin, Bing Xue, Bingxuan Wang, Bingzheng Xu,
  Bochao Wu, Bowei Zhang, Chaofan Lin, Chen Dong, and 1 others. 2025.
\newblock Deepseek-v3. 2: Pushing the frontier of open large language models.
\newblock \emph{arXiv preprint arXiv:2512.02556}.

\bibitem[{Liu et~al.(2024)Liu, Du, Xiang, Wang, and Luo}]{liu2024long}
Siyuan Liu, Jiawei Du, Sicheng Xiang, Zibo Wang, and Dingsheng Luo. 2024.
\newblock Long-horizon embodied planning with implicit logical inference and
  hallucination mitigation.
\newblock \emph{arXiv preprint arXiv:2409.15658}.

\bibitem[{Liu et~al.(2026)Liu, Sun, Chao, Yang, Xie, Zhang, Li, Tang, Chen,
  Zhang et~al.}]{liu2026checkvla}
Yushan Liu, Peibo Sun, Xintao Chao, Zhenyang Yang, Yifan Xie, Lingfeng Zhang,
  Shoujie Li, Chenyu Tang, Fang Chen, Xiao-Ping Zhang, and 1 others. 2026.
\newblock Checkvla: Execution-time verification with action-conditioned world
  model for long-horizon mobile manipulation.
\newblock \emph{arXiv preprint arXiv:2607.26789}.

\bibitem[{Molinari and Ciravegna(2025)}]{molinari2025reason}
Gianni Molinari and Fabio Ciravegna. 2025.
\newblock Reason-plan-react: A reasoner-planner supervising a react executor
  for complex enterprise tasks.
\newblock \emph{arXiv preprint arXiv:2512.03560}.

\bibitem[{Petruzzellis et~al.(2025)Petruzzellis, Cornelio, and
  Lio}]{petruzzellis2025hierarchical}
Flavio Petruzzellis, Cristina Cornelio, and Pietro Lio. 2025.
\newblock \href {https://proceedings.mlr.press/v267/petruzzellis25a.html}
  {Hierarchical planning for complex tasks with knowledge graph-{RAG} and
  symbolic verification}.
\newblock In \emph{Proceedings of the 42nd International Conference on Machine
  Learning}, volume 267 of \emph{Proceedings of Machine Learning Research},
  pages 49105--49127. PMLR.

\bibitem[{Reimers and Gurevych(2019)}]{reimers-2019-sentence-bert}
Nils Reimers and Iryna Gurevych. 2019.
\newblock \href {https://arxiv.org/abs/1908.10084} {Sentence-bert: Sentence
  embeddings using siamese bert-networks}.
\newblock In \emph{Proceedings of the 2019 Conference on Empirical Methods in
  Natural Language Processing}. Association for Computational Linguistics.

\bibitem[{Shinn et~al.(2023)Shinn, Cassano, Gopinath, Narasimhan, and
  Yao}]{shinn2023reflexion}
Noah Shinn, Federico Cassano, Ashwin Gopinath, Karthik Narasimhan, and Shunyu
  Yao. 2023.
\newblock \href {https://doi.org/10.52202/075280-0377} {Reflexion: Language
  agents with verbal reinforcement learning}.
\newblock In \emph{Advances in Neural Information Processing Systems},
  volume~36, pages 8634--8652. Curran Associates, Inc.

\bibitem[{Shridhar et~al.(2021)Shridhar, Yuan, C\^ot\'e, Bisk, Trischler, and
  Hausknecht}]{ALFWorld20}
Mohit Shridhar, Xingdi Yuan, Marc-Alexandre C\^ot\'e, Yonatan Bisk, Adam
  Trischler, and Matthew Hausknecht. 2021.
\newblock \href {https://arxiv.org/abs/2010.03768} {{ALFWorld}: Aligning text
  and embodied environments for interactive learning}.
\newblock In \emph{Proceedings of the International Conference on Learning
  Representations}.

\bibitem[{Singh et~al.(2022)Singh, Blukis, Mousavian, Goyal, Xu, Tremblay, Fox,
  Thomason, and Garg}]{singh2022progprompt}
Ishika Singh, Valts Blukis, Arsalan Mousavian, Ankit Goyal, Danfei Xu, Jonathan
  Tremblay, Dieter Fox, Jesse Thomason, and Animesh Garg. 2022.
\newblock \href {https://openreview.net/forum?id=3K4-U_5cRw} {Progprompt:
  Generating situated robot task plans using large language models}.
\newblock In \emph{Workshop on Language and Robotics at CoRL 2022}.

\bibitem[{Sun et~al.(2023)Sun, Zhuang, Kong, Dai, and
  Zhang}]{sun2023adaplanner}
Haotian Sun, Yuchen Zhuang, Lingkai Kong, Bo~Dai, and Chao Zhang. 2023.
\newblock \href {https://doi.org/10.52202/075280-2537} {Adaplanner: Adaptive
  planning from feedback with language models}.
\newblock In \emph{Advances in Neural Information Processing Systems},
  volume~36, pages 58202--58245. Curran Associates, Inc.

\bibitem[{Wang et~al.(2023)Wang, Xu, Lan, Hu, Lan, Lee, and
  Lim}]{wang-etal-2023-plan}
Lei Wang, Wanyu Xu, Yihuai Lan, Zhiqiang Hu, Yunshi Lan, Roy Ka-Wei Lee, and
  Ee-Peng Lim. 2023.
\newblock \href {https://doi.org/10.18653/v1/2023.acl-long.147} {Plan-and-solve
  prompting: Improving zero-shot chain-of-thought reasoning by large language
  models}.
\newblock In \emph{Proceedings of the 61st Annual Meeting of the Association
  for Computational Linguistics (Volume 1: Long Papers)}, pages 2609--2634,
  Toronto, Canada. Association for Computational Linguistics.

\bibitem[{Wang et~al.(2022)Wang, Jansen, C{\^o}t{\'e}, and
  Ammanabrolu}]{wang-etal-2022-scienceworld}
Ruoyao Wang, Peter Jansen, Marc-Alexandre C{\^o}t{\'e}, and Prithviraj
  Ammanabrolu. 2022.
\newblock \href {https://doi.org/10.18653/v1/2022.emnlp-main.775}
  {{S}cience{W}orld: Is your agent smarter than a 5th grader?}
\newblock In \emph{Proceedings of the 2022 Conference on Empirical Methods in
  Natural Language Processing}, pages 11279--11298, Abu Dhabi, United Arab
  Emirates. Association for Computational Linguistics.

\bibitem[{Wu et~al.(2025)Wu, Luo, Qu, and Wang}]{wu2025meta}
Chunlong Wu, Ye~Luo, Zhibo Qu, and Min Wang. 2025.
\newblock Meta-policy reflexion: Reusable reflective memory and rule
  admissibility for resource-efficient llm agent.
\newblock \emph{arXiv preprint arXiv:2509.03990}.

\bibitem[{Xiong et~al.(2025)Xiong, Song, Dong, Zhao, Song, XWang, and
  Li}]{xiong-etal-2025-mpo}
Weimin Xiong, Yifan Song, Qingxiu Dong, Bingchan Zhao, Feifan Song, XWang, and
  Sujian Li. 2025.
\newblock \href {https://doi.org/10.18653/v1/2025.findings-emnlp.210} {{MPO}:
  Boosting {LLM} agents with meta plan optimization}.
\newblock In \emph{Findings of the Association for Computational Linguistics:
  EMNLP 2025}, pages 3914--3935, Suzhou, China. Association for Computational
  Linguistics.

\bibitem[{Xu et~al.(2023)Xu, Peng, Lei, Mukherjee, Liu, and Xu}]{ReWoo2023}
Binfeng Xu, Zhiyuan Peng, Bowen Lei, Subhabrata Mukherjee, Yuchen Liu, and
  Dongkuan Xu. 2023.
\newblock \href {https://doi.org/10.48550/arXiv.2305.18323} {Rewoo: Decoupling
  reasoning from observations for efficient augmented language models}.
\newblock \emph{CoRR}, abs/2305.18323.

\bibitem[{Yang et~al.(2025)Yang, Chen, Zhang, Zhao, Qian, Wang, Wang,
  Koripella, Movahedi, Li, Ji, Zhang, and Zhang}]{yang2025embodiedbench}
Rui Yang, Hanyang Chen, Junyu Zhang, Mark Zhao, Cheng Qian, Kangrui Wang,
  Qineng Wang, Teja~Venkat Koripella, Marziyeh Movahedi, Manling Li, Heng Ji,
  Huan Zhang, and Tong Zhang. 2025.
\newblock \href {https://proceedings.mlr.press/v267/yang25f.html}
  {{E}mbodied{B}ench: Comprehensive benchmarking multi-modal large language
  models for vision-driven embodied agents}.
\newblock In \emph{Proceedings of the 42nd International Conference on Machine
  Learning}, volume 267 of \emph{Proceedings of Machine Learning Research},
  pages 70576--70631. PMLR.

\bibitem[{Yang et~al.(2026)Yang, Jin, Zhao, Wu, Zhou, Wang, Wang, Zhou, and
  Metaxas}]{yang2026space}
Yanting Yang, Can Jin, Jinman Zhao, Jiahao Wu, Yang Zhou, Zhepeng Wang,
  Zhendong Wang, Mu~Zhou, and Dimitris~N Metaxas. 2026.
\newblock Act more, decide less: Skill-guided adaptive action chunking for
  long-horizon llm agents.
\newblock \emph{arXiv preprint arXiv:2609.02042}.

\bibitem[{Yao et~al.(2022)Yao, Zhao, Yu, Du, Shafran, Narasimhan, and
  Cao}]{yao2022react}
Shunyu Yao, Jeffrey Zhao, Dian Yu, Nan Du, Izhak Shafran, Karthik~R Narasimhan,
  and Yuan Cao. 2022.
\newblock React: Synergizing reasoning and acting in language models.
\newblock In \emph{Proceedings of the International Conference on Learning
  Representations}.

\end{thebibliography}

\appendix
% Keep appendix floats out of the preceding bibliography column.
\suppressfloats[t]

% Keep appendix text flowing around top floats.
\begingroup
\raggedbottom
% Advance past both columns while keeping queued tables at the top.
\makeatletter
\newcommand{\appendixpagebreak}{\par\null\newpage\if@firstcolumn\else\null\newpage\fi}
\makeatother

\section{Token Accounting and Efficiency Analysis}
\label{sec:appendix_token_accounting}

We measure token cost by weighting input and output token counts according to their relative prices and summing the contributions.
Table~\ref{tab:token_weight} lists these weights, with the output-token weight normalized to 1.

% Queue Figure 5 before Table 6 so the two top floats appear in that order.
\begin{figure}[t]
    \centering
    \includegraphics[width=\columnwidth]{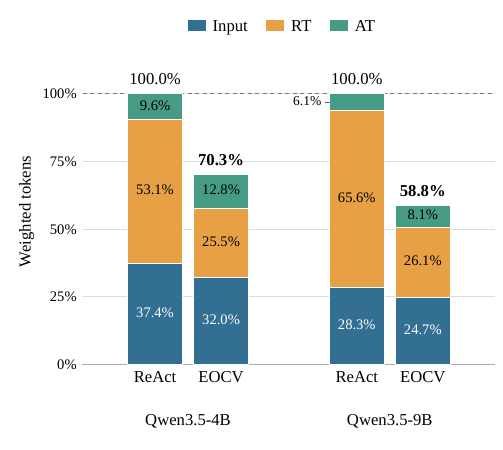}
    \caption{ALFWorld token-cost decomposition without caching. Bar segment heights show price-weighted token costs, not raw token counts. Input, RT, and AT have weights 0.125, 1, and 1. Each model's ReAct total cost is normalized to 100\%.}
    \label{fig:token_efficiency}
\end{figure}

\begin{table}[tp]
\centering
\small
\setlength{\tabcolsep}{4pt}
\begin{tabular*}{\columnwidth}{@{\extracolsep{\fill}}lcc@{}}
\toprule
\textbf{Model} & \textbf{Input Weight} & \textbf{Output Weight} \\
\midrule
Qwen3.5-4B/9B & $1/8$ & $1$ \\
GPT-5.5 & $1/6$ & $1$ \\
DeepSeek-V4.1-Flash & $1/4$ & $1$ \\
DeepSeek-V3.2 & $2/3$ & $1$ \\
Qwen3.6-Plus/Flash & $1/6$ & $1$ \\
\bottomrule
\end{tabular*}
\caption{Token weights used to compute weighted token usage. The output-token weight is normalized to 1; each input-token weight is the model's input-to-output price ratio.}
\label{tab:token_weight}
\end{table}

\subsection{Sources of Token Savings}
\label{sec:appendix_efficiency_analysis}

EOCV gains efficiency by allowing multiple actions to share one executor call.
A reactive executor generates one action per call, repeatedly receiving the task instructions and interaction history and, when reasoning is enabled, starting a new reasoning phase.
EOCV instead generates up to five actions per call and executes them until the batch ends or replanning is needed.
Actions within a batch share the same context prefill and internal reasoning, reducing repeated input and potentially reducing reasoning-token generation.

Batching also adds output costs: EOCV generates expectations, and some proposed actions may be discarded when execution stops early.
These tokens are counted even when the actions are never executed.
EOCV therefore gains a net advantage when input and reasoning savings outweigh the additional action-response output and any model-based verification cost.

\begin{figure}[t]
    \centering
    \includegraphics[width=\columnwidth]{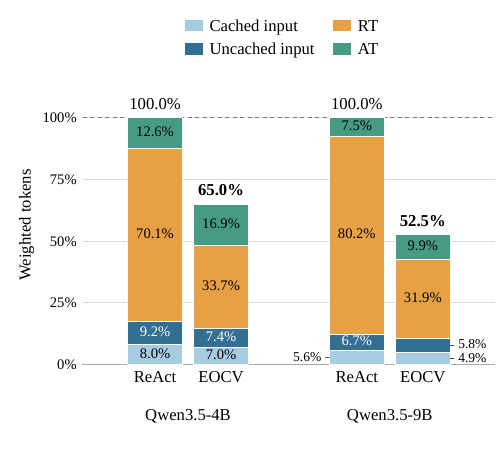}
    \caption{ALFWorld token-cost decomposition with caching. Bar segment heights show price-weighted token costs, not raw token counts. Cached, uncached Input, RT, and AT have weights 0.025, 0.125, 1, and 1. Each model's ReAct total cost is normalized to 100}
    \label{fig:token_efficiency_cache}
\end{figure}

\subsection{ALFWorld Case Study}

Figure~\ref{fig:token_efficiency} compares weighted token usage of executor-only ReAct and EOCV runs using Qwen3.5-4B and Qwen3.5-9B on the same 134 ALFWorld tasks.
We separate output into reported internal reasoning tokens (RT) and all remaining action-response tokens (AT), including visible thoughts, commands, expectations, formatting, and unexecuted suffixes.
In both pairs, EOCV's savings in input and RT costs outweigh its increased AT costs, reducing weighted usage to $0.703\times$ ReAct for 4B and $0.588\times$ for 9B.
RT contributes most of the savings.

\subsection{Effect of Prefix Caching}

Prefix caching reuses stored key--value states for matching input prefixes, reducing repeated prefill computation and the price of cached input tokens.
We recompute token costs for the same runs by assigning weights of 0.025 and 0.125 to cached and uncached input tokens, respectively, while keeping output-token weights unchanged.

Figure~\ref{fig:token_efficiency_cache} gives EOCV costs of $0.650\times$ and $0.525\times$ ReAct for 4B and 9B.
Caching discounts input but leaves reasoning-token generation costs unchanged, making RT a larger share of ReAct's total cost.
EOCV's RT savings therefore remain valuable, and its relative cost advantage increases in these comparisons.
% This change reflects input discounts and renormalization, rather than a further reduction in generated tokens.

\section{Additional Experimental Results}
\label{sec:appendix_result}
This appendix complements the main experimental section with detailed results and analysis omitted from the main text.

\subsection{Expectation Skipping}
\label{sec:appendix_skip_verify}

The main experiments use a predefined rule set to skip expectation generation and verification for information-seeking actions.
The actions themselves are still executed.
For the automatic variant, the executor is prompted at each step to output an \texttt{action-is-skippable} tag.
When this tag is emitted, EOCV skips both expectation prediction and verification for that action.
Table~\ref{tab:skip_strategy} reports the precision and recall of the automatic tag, together with the downstream reward and token usage when the tag replaces the predefined rule set.

\begin{table}[tp]
\centering
\small
\setlength{\tabcolsep}{3pt}
\begin{tabular*}{\columnwidth}{@{\extracolsep{\fill}}lrrrr@{}}
\toprule
\textbf{Strategy} & \textbf{Prec.} & \textbf{Rec.} & \textbf{Reward} & \textbf{Tokens} \\
\midrule
Predefined & -- & -- & 94.78 & 1.00$\times$ \\
Automatic & 99.68 & 99.36 & 94.03 & 0.92$\times$ \\
\bottomrule
\end{tabular*}
\caption{Predefined and automatic expectation skipping on ALFWorld with DeepSeek-V3.2. Precision (Prec.) and recall (Rec.) are computed on 400 labeled actions for the automatic \texttt{action-is-skippable} tag. Token usage is normalized to predefined skipping; dashes indicate that a metric is not applicable.}
\label{tab:skip_strategy}
\end{table}

\paragraph{Skipping Alone.}
We further evaluate a skip-only ablation on ALFWorld with DeepSeek-V3.2.
It executes information-seeking actions without expectation generation or verification, but limits each batch to at most one non-information-seeking action and terminates the batch immediately after executing it.
% User-supplied skip-only result (2026-10-13): average=0.9477611940298507,
% input_tokens=4462361, output_tokens=260036, cached_tokens=0,
% verify_input_tokens=0, verify_output_tokens=0; weighted tokens=(2/3)*input+output.
It achieves a reward of 94.78, matching both Reactive and full EOCV, while using 3.23M weighted tokens, compared with 3.79M for Reactive and 2.45M for full EOCV (Table~\ref{tab:primary_executor_details}).
This amounts to approximately 15\% fewer tokens than Reactive but 32\% more than full EOCV.
Skipping alone therefore cannot recover the token savings enabled by verified batched execution.

\subsection{Verification Outcomes}
\label{sec:appendix_verification_outcomes}

Table~\ref{tab:skip_verify_distribution} reports three verification outcomes:
\textbf{Skip}, where expectation generation and verification are bypassed for information-seeking actions;
\textbf{Pass}, where the expectation is judged consistent with the observation;
and \textbf{Fail}, where it is judged inconsistent.
Actions without a recorded verification outcome are excluded.

Skipping accounts for a substantial share of actions across settings, reducing verification overhead.
For verified actions, EOCV continues execution after a pass and interrupts it after a failure to limit error propagation.

% Qwen3.5, GPT-5.5, and DeepSeek-V4.1-Flash: outputs/skip_replan_rate/results.json,
% generated by scripts/calc_skip_replan_rate.py from config/experiment_results.toml.
% Unclassified action records excluded: 4B 118, 9B 120, GPT-5.5 255, DeepSeek 212.
% EB-Habitat: latest user-supplied counts (Skip/Pass/Fail), 2026-10-12.
% Qwen3.6-Flash: 2988/494/1463; Qwen3.6-Plus: 2437/567/1225.
% These replace the previously reported Qwen3.6-Flash percentages.
\begin{table}[tp]
\centering
\small
\setlength{\tabcolsep}{3pt}
\begin{tabular*}{\columnwidth}{@{\extracolsep{\fill}}lrrr@{}}
\toprule
\textbf{Backbone} & \textbf{Skip} & \textbf{Pass} & \textbf{Fail} \\
\midrule
\multicolumn{4}{l}{\textbf{ALFWorld}} \\
DeepSeek-V3.2 & 75.55 & 11.60 & 12.85 \\
Qwen3.5-4B & 80.39 & 8.64 & 10.96 \\
Qwen3.5-9B & 77.00 & 10.78 & 12.23 \\
\midrule
\multicolumn{4}{l}{\textbf{ScienceWorld}} \\
DeepSeek-V3.2 & 49.56 & 33.06 & 17.38 \\
GPT-5.5 & 45.65 & 44.44 & 9.91 \\
DeepSeek-V4.1-Flash & 55.73 & 35.81 & 8.46 \\
\midrule
\multicolumn{4}{l}{\textbf{EB-Habitat}} \\
Qwen3.6-Flash & 60.42 & 9.99 & 29.59 \\
Qwen3.6-Plus & 57.63 & 13.41 & 28.97 \\
\bottomrule
\end{tabular*}
\caption{Distribution (\%) of EOCV execution outcomes. Skip denotes actions for which expectation prediction and verification are omitted.}
\label{tab:skip_verify_distribution}
\end{table}

\subsection{Verifier Difficulty Analysis}
\label{sec:appendix_verifier_difficulty}

We analyze the same 300 randomly sampled, manually labeled expectation--observation pairs used in the EB-Habitat precision and recall evaluation in Table~\ref{tab:verifier_analysis}.
Each sample includes an action, its expectation, and the resulting observation from a Qwen3.6-Flash execution trajectory; the same model also serves as the verifier.
We use action execution outcomes as a proxy for verification difficulty.
We classify failed actions as Easy cases because they leave the environment state unchanged, making the absence of an expected effect relatively easy to identify.
Successful actions change the environment state, but execution success alone does not guarantee that the expected effect has occurred.
These Hard cases require the verifier to interpret changes in object states or relations and determine whether they match the expectation, placing greater demands on its reasoning.

We define accuracy within each group as the fraction of verifier decisions that agree with the manually assigned consistency labels.
The verifier correctly classifies 202 of 232 Easy cases and 44 of 68 Hard cases, giving accuracies of 87.07\% and 64.71\%, respectively.

\subsection{Hyperparameter Sensitivity}
\label{sec:appendix_hyperparameters}
These analyses use ALFWorld with DeepSeek-V3.2.
Given ALFWorld's binary rewards, we report mean reward and weighted token usage
across four independent runs for each ablation setting to obtain more stable estimates.

\begin{figure}[t]
    \centering
    \includegraphics[width=0.9\columnwidth]{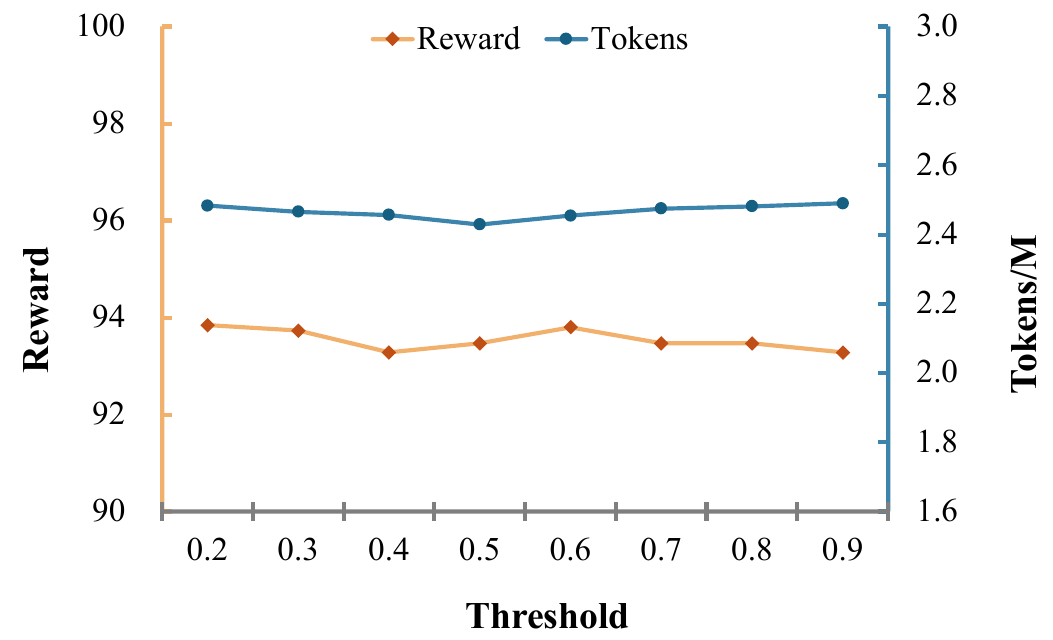}
    \vspace{-0.7em}
    \caption{Analysis of verification threshold $\tau$. Token usage is measured in millions (M).}
    \label{fig:threshold}
\end{figure}

\paragraph{Verification Threshold.}
\label{sec:threshold_analysis}
Figure~\ref{fig:threshold} studies how the verification threshold $\tau$ in expectation--observation verification affects the efficiency--reward trade-off.
Across a broad range of thresholds, average reward remains largely stable, indicating that the verification criterion is not overly sensitive and does not require fine-grained tuning.
Token consumption exhibits a slight U-shaped trend: overly permissive thresholds admit more inconsistent steps, increasing downstream correction cost, whereas overly strict thresholds trigger frequent early termination and regeneration, also increasing token usage.
This stability motivates our use of an intermediate threshold.

\begin{figure}[t]
    \centering
    \includegraphics[width=0.9\columnwidth]{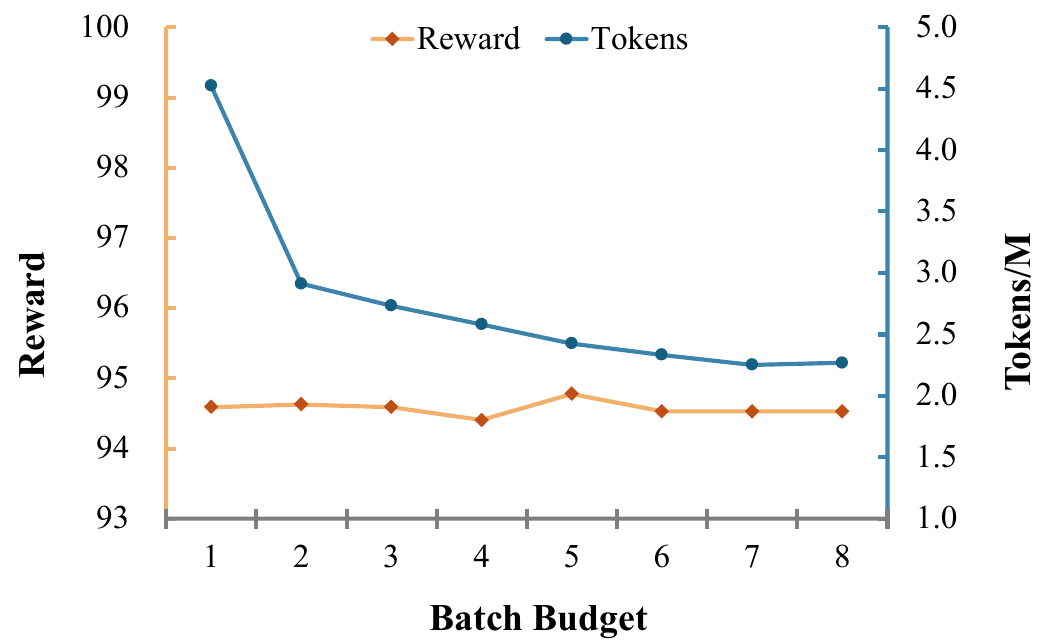}
    \vspace{-0.7em}
    \caption{Analysis of batch budget $B$. Token usage is measured in millions (M).}
    \label{fig:batch_budget}
\end{figure}

\paragraph{Batch Budget.}
Figure~\ref{fig:batch_budget} examines the impact of the batch budget $B$.
As $B$ increases, token usage consistently decreases, reflecting improved efficiency from generating more actions within a single model invocation.
At the same time, average reward remains largely stable across a broad range of $B$.
When $B$ becomes large, efficiency gains gradually saturate while reward shows only minor fluctuations, suggesting diminishing returns from further increasing the batch budget.
We therefore use a moderate budget, which captures most of the efficiency gain before the saturation regime.

\subsection{EOCV Results}
\label{sec:appendix_text_and_multimodal_results}

\subsubsection{Hypothesis Testing}
\label{sec:appendix_reward_summary}
We compare EOCV with Reactive across 28 configurations matched by benchmark, backbone, and controller setting.
ALFWorld and ScienceWorld each contribute 12 pairs: three backbones evaluated without a controller or with ReflAct, Reflexion, or MPO.
EB-Habitat contributes four pairs: Qwen3.6-Flash and Qwen3.6-Plus without controllers, plus Qwen3.6-Flash with ReflAct or Reflexion.

For configuration $i$, let $d_i = r_i^{\mathrm{EOCV}} - r_i^{\mathrm{Reactive}}$ on the 0--100 reward scale, with each pair weighted equally irrespective of task count.
A two-sided paired $t$-test of $H_0:\mathbb{E}[d_i]=0$ at $\alpha=0.05$ gives a mean difference of $+0.0458$ points (95\% paired-$t$ confidence interval: [$-0.3406$, $+0.4321$]), with $t(27)=0.2430$ ($p=0.8098$).
This analysis provides an exploratory assessment of the average reward difference across the evaluated configurations, whose shared backbones and evaluation tasks may induce dependence.

\subsubsection{Additional Results}
\label{sec:appendix_additional_results}

Table~\ref{tab:primary_executor_details} compares Reactive, Batched, and EOCV executors without controllers across all three benchmarks.
Table~\ref{tab:ebhabitat_details} further breaks down EB-Habitat results by task category and vision-language backbone.
For EB-Habitat, rewards are averaged across the six categories, while token totals are summed across categories and include vision-language verification costs; token ratios use these totals.

% Generated by paper/scripts/update_appendix_results.py from paper/raw_results/*.csv.
\begin{table*}[tp]
\centering
\small
\setlength{\tabcolsep}{3pt}
\begin{tabular*}{\textwidth}{@{\extracolsep{\fill}}ll*{7}{r}@{}}
\toprule
\multirow{2}{*}{\textbf{Backbone}} & \multirow{2}{*}{\textbf{Executor}} & \multirow{2}{*}{\textbf{Reward}} & \textbf{Input} & \textbf{Output} & \multicolumn{2}{c}{\textbf{Weighted Tokens}} & \multicolumn{2}{c}{\textbf{Time}} \\
\cmidrule(lr){6-7}\cmidrule(lr){8-9}
 & & & \textbf{(M)} & \textbf{(M)} & \textbf{(M)} & \textbf{Ratio} & \textbf{(s)} & \textbf{Ratio} \\
\midrule
\multicolumn{9}{l}{\textbf{ALFWorld}} \\
\addlinespace[2pt]
\multirow{3}{*}{Qwen3.5-4B} & Reactive & 79.85 & 7.15 & 1.50 & 2.39 & 1.00$\times$ & 17\,394.41 & 1.00$\times$ \\
 & Batched & 73.88 & 5.44 & 1.03 & 1.71 & 0.71$\times$ & 9\,603.92 & 0.55$\times$ \\
 & Ours & 79.10 & 6.12 & 0.92 & 1.68 & 0.70$\times$ & 11\,170.33 & 0.64$\times$ \\
\addlinespace[3pt]
\multirow{3}{*}{Qwen3.5-9B} & Reactive & 91.79 & 6.15 & 1.95 & 2.72 & 1.00$\times$ & 25\,984.66 & 1.00$\times$ \\
 & Batched & 86.57 & 4.56 & 0.87 & 1.44 & 0.53$\times$ & 12\,973.72 & 0.50$\times$ \\
 & Ours & 91.79 & 5.36 & 0.93 & 1.60 & 0.59$\times$ & 13\,804.76 & 0.53$\times$ \\
\addlinespace[3pt]
\multirow{3}{*}{DeepSeek-V3.2} & Reactive & 94.78 & 5.52 & 0.11 & 3.79 & 1.00$\times$ & -- & -- \\
 & Batched & 90.30 & 2.39 & 0.13 & 1.73 & 0.46$\times$ & -- & -- \\
 & Ours & 94.78 & 3.31 & 0.24 & 2.45 & 0.65$\times$ & -- & -- \\
\midrule
\multicolumn{9}{l}{\textbf{ScienceWorld}} \\
\addlinespace[2pt]
\multirow{3}{*}{GPT-5.5} & Reactive & 94.30 & 27.17 & 0.48 & 5.01 & 1.00$\times$ & -- & -- \\
 & Batched & 92.88 & 11.42 & 0.36 & 2.27 & 0.45$\times$ & -- & -- \\
 & Ours & 94.81 & 15.69 & 0.50 & 3.11 & 0.62$\times$ & -- & -- \\
\addlinespace[3pt]
\multirow{3}{*}{DS-V4.1-Flash} & Reactive & 86.46 & 32.45 & 0.97 & 9.08 & 1.00$\times$ & -- & -- \\
 & Batched & 76.00 & 10.97 & 1.23 & 3.98 & 0.44$\times$ & -- & -- \\
 & Ours & 86.39 & 21.11 & 1.06 & 6.34 & 0.70$\times$ & -- & -- \\
\addlinespace[3pt]
\multirow{3}{*}{DeepSeek-V3.2} & Reactive & 87.36 & 29.81 & 0.55 & 20.42 & 1.00$\times$ & -- & -- \\
 & Batched & 78.80 & 9.68 & 0.53 & 6.98 & 0.34$\times$ & -- & -- \\
 & Ours & 85.42 & 20.99 & 1.06 & 15.06 & 0.74$\times$ & -- & -- \\
\midrule
\multicolumn{9}{l}{\textbf{EB-Habitat}} \\
\addlinespace[2pt]
\multirow{3}{*}{Qwen3.6-Flash} & Reactive & 58.10 & 28.57 & 1.66 & 6.42 & 1.00$\times$ & -- & -- \\
 & Batched & 48.88 & 7.41 & 0.62 & 1.85 & 0.29$\times$ & -- & -- \\
 & Ours & 59.49 & 13.73 & 1.78 & 4.06 & 0.63$\times$ & -- & -- \\
\addlinespace[3pt]
\multirow{3}{*}{Qwen3.6-Plus} & Reactive & 75.78 & 21.13 & 1.34 & 4.87 & 1.00$\times$ & -- & -- \\
 & Batched & 64.24 & 6.67 & 1.17 & 2.28 & 0.47$\times$ & -- & -- \\
 & Ours & 76.04 & 12.31 & 1.78 & 3.84 & 0.79$\times$ & -- & -- \\
\bottomrule
\end{tabular*}
\caption{Executor results by benchmark and backbone. Reward is on $[0,100]$; token totals are in millions (M). Qwen3.5 times are summed task durations in seconds. Ratios are relative to Reactive within each backbone. EB-Habitat totals sum the six categories in Table~\ref{tab:ebhabitat_details}. Dashes denote unavailable values. DS-V4.1-Flash denotes DeepSeek-V4.1-Flash.}
\label{tab:primary_executor_details}
\label{tab:appendix_baseline_token_usage}
\end{table*}

\begin{table*}[tp]
\centering
\small
\setlength{\tabcolsep}{3pt}
\begin{tabular*}{\textwidth}{@{\extracolsep{\fill}}l*{8}{r}@{}}
\toprule
\multirow{2}{*}{\textbf{Category}} & \multicolumn{2}{c}{\textbf{Reactive}} & \multicolumn{3}{c}{\textbf{Batched}} & \multicolumn{3}{c}{\textbf{Ours}} \\
\cmidrule(lr){2-3}\cmidrule(lr){4-6}\cmidrule(lr){7-9}
 & \textbf{Reward} & \textbf{Tokens (M)} & \textbf{Reward} & \textbf{Tokens (M)} & \textbf{Ratio} & \textbf{Reward} & \textbf{Tokens (M)} & \textbf{Ratio} \\
\midrule
\multicolumn{9}{l}{\textbf{Qwen3.6-Flash}} \\
\addlinespace[2pt]
Base & 87.25 & 0.56 & 79.75 & 0.14 & 0.25$\times$ & 89.75 & 0.27 & 0.49$\times$ \\
Common Sense & 44.00 & 1.17 & 31.47 & 0.47 & 0.40$\times$ & 45.00 & 0.69 & 0.59$\times$ \\
Complex Instr. & 54.00 & 1.06 & 52.07 & 0.30 & 0.28$\times$ & 56.00 & 0.61 & 0.58$\times$ \\
Long Horizon & 33.42 & 1.36 & 19.71 & 0.32 & 0.24$\times$ & 34.50 & 0.74 & 0.55$\times$ \\
Spatial Rel. & 68.00 & 1.23 & 65.83 & 0.33 & 0.26$\times$ & 70.67 & 0.91 & 0.73$\times$ \\
Visual App. & 61.92 & 1.05 & 44.48 & 0.30 & 0.29$\times$ & 61.00 & 0.84 & 0.80$\times$ \\
\cmidrule(lr){1-9}
Average & 58.10 & 1.07 & 48.88 & 0.31 & 0.29$\times$ & 59.49 & 0.68 & 0.63$\times$ \\
\midrule
\multicolumn{9}{l}{\textbf{Qwen3.6-Plus}} \\
\addlinespace[2pt]
Base & 96.00 & 0.31 & 88.75 & 0.15 & 0.47$\times$ & 95.50 & 0.25 & 0.81$\times$ \\
Common Sense & 72.50 & 0.87 & 54.00 & 0.45 & 0.51$\times$ & 71.50 & 0.63 & 0.73$\times$ \\
Complex Instr. & 84.00 & 0.76 & 73.50 & 0.38 & 0.50$\times$ & 87.50 & 0.64 & 0.84$\times$ \\
Long Horizon & 58.33 & 0.98 & 42.17 & 0.36 & 0.36$\times$ & 59.00 & 0.64 & 0.66$\times$ \\
Spatial Rel. & 66.17 & 1.22 & 60.75 & 0.69 & 0.56$\times$ & 64.25 & 1.08 & 0.89$\times$ \\
Visual App. & 77.67 & 0.73 & 66.25 & 0.27 & 0.37$\times$ & 78.50 & 0.58 & 0.80$\times$ \\
\cmidrule(lr){1-9}
Average & 75.78 & 0.81 & 64.24 & 0.38 & 0.47$\times$ & 76.04 & 0.64 & 0.79$\times$ \\
\bottomrule
\end{tabular*}
\caption{Detailed EB-Habitat results by task category and vision-language backbone. Rewards are rescaled to $[0,100]$; tokens are price-weighted totals in millions (M). Ratios are relative to Reactive within each category. Average rows report category means; their token ratios divide each mean by the corresponding Reactive mean.}
\label{tab:ebhabitat_details}
\end{table*}

Table~\ref{tab:controller_composition} pairs Reactive and EOCV executors with ReflAct, MPO, and Reflexion on each text benchmark, plus ReflAct and Reflexion on EB-Habitat.

% Generated by paper/scripts/update_appendix_results.py from paper/raw_results/*.csv.
\begin{table*}[!t]
\centering
\small
\setlength{\tabcolsep}{3pt}
\renewcommand{\arraystretch}{0.94}
\begin{tabular*}{\textwidth}{@{\extracolsep{\fill}}ll*{7}{r}@{}}
\toprule
\multirow{2}{*}{\textbf{Controller}} & \multirow{2}{*}{\textbf{Executor}} & \multirow{2}{*}{\textbf{Reward}} & \textbf{Input} & \textbf{Output} & \multicolumn{2}{c}{\textbf{Weighted Tokens}} & \multicolumn{2}{c}{\textbf{Time}} \\
\cmidrule(lr){6-7}\cmidrule(lr){8-9}
 & & & \textbf{(M)} & \textbf{(M)} & \textbf{(M)} & \textbf{Ratio} & \textbf{(s)} & \textbf{Ratio} \\
\specialrule{0.8pt}{3pt}{2pt}
\multicolumn{9}{l}{\textbf{ALFWorld}} \\
\multicolumn{9}{l}{\textit{Qwen3.5-4B}} \\
\addlinespace[1pt]
\multirow{2}{*}{ReflAct} & Reactive & 81.34 & 7.89 & 3.48 & 4.46 & 1.00$\times$ & 40\,540.94 & 1.00$\times$ \\
 & Ours & 81.34 & 5.45 & 2.22 & 2.90 & 0.65$\times$ & 25\,778.94 & 0.64$\times$ \\
\addlinespace[1pt]
\multirow{2}{*}{MPO} & Reactive & 88.06 & 5.11 & 1.54 & 2.18 & 1.00$\times$ & 14\,173.30 & 1.00$\times$ \\
 & Ours & 89.55 & 6.18 & 1.02 & 1.79 & 0.82$\times$ & 9\,647.12 & 0.68$\times$ \\
\addlinespace[1pt]
\multirow{2}{*}{Reflexion} & Reactive & 90.30 & 13.59 & 3.97 & 5.67 & 1.00$\times$ & 36\,263.58 & 1.00$\times$ \\
 & Ours & 88.81 & 12.78 & 2.35 & 3.95 & 0.70$\times$ & 22\,008.90 & 0.61$\times$ \\
\specialrule{0.3pt}{2pt}{1pt}
\multicolumn{9}{l}{\textit{Qwen3.5-9B}} \\
\addlinespace[1pt]
\multirow{2}{*}{ReflAct} & Reactive & 91.04 & 7.58 & 1.51 & 2.46 & 1.00$\times$ & 22\,534.70 & 1.00$\times$ \\
 & Ours & 91.04 & 4.88 & 0.90 & 1.51 & 0.62$\times$ & 13\,605.20 & 0.60$\times$ \\
\addlinespace[1pt]
\multirow{2}{*}{MPO} & Reactive & 90.30 & 6.03 & 2.38 & 3.13 & 1.00$\times$ & 34\,745.77 & 1.00$\times$ \\
 & Ours & 89.55 & 4.89 & 1.37 & 1.98 & 0.63$\times$ & 20\,177.39 & 0.58$\times$ \\
\addlinespace[1pt]
\multirow{2}{*}{Reflexion} & Reactive & 98.51 & 7.99 & 2.75 & 3.75 & 1.00$\times$ & 36\,062.00 & 1.00$\times$ \\
 & Ours & 97.01 & 9.36 & 2.11 & 3.28 & 0.87$\times$ & 28\,296.39 & 0.78$\times$ \\
\specialrule{0.3pt}{2pt}{1pt}
\multicolumn{9}{l}{\textit{DeepSeek-V3.2}} \\
\addlinespace[1pt]
\multirow{2}{*}{ReflAct} & Reactive & 88.81 & 4.76 & 0.07 & 3.25 & 1.00$\times$ & -- & -- \\
 & Ours & 87.31 & 3.13 & 0.19 & 2.28 & 0.70$\times$ & -- & -- \\
\addlinespace[1pt]
\multirow{2}{*}{MPO} & Reactive & 94.78 & 3.50 & 0.06 & 2.40 & 1.00$\times$ & -- & -- \\
 & Ours & 95.52 & 2.35 & 0.15 & 1.72 & 0.72$\times$ & -- & -- \\
\addlinespace[1pt]
\multirow{2}{*}{Reflexion} & Reactive & 98.51 & 6.04 & 0.10 & 4.13 & 1.00$\times$ & -- & -- \\
 & Ours & 99.25 & 3.59 & 0.27 & 2.66 & 0.64$\times$ & -- & -- \\
\specialrule{0.8pt}{3pt}{2pt}
\multicolumn{9}{l}{\textbf{ScienceWorld}} \\
\multicolumn{9}{l}{\textit{GPT-5.5}} \\
\addlinespace[1pt]
\multirow{2}{*}{ReflAct} & Reactive & 94.18 & 18.91 & 0.68 & 3.83 & 1.00$\times$ & -- & -- \\
 & Ours & 95.35 & 9.29 & 0.51 & 2.06 & 0.54$\times$ & -- & -- \\
\addlinespace[1pt]
\multirow{2}{*}{MPO} & Reactive & 93.20 & 12.45 & 0.36 & 2.44 & 1.00$\times$ & -- & -- \\
 & Ours & 94.44 & 6.02 & 0.33 & 1.34 & 0.55$\times$ & -- & -- \\
\addlinespace[1pt]
\multirow{2}{*}{Reflexion} & Reactive & 97.18 & 20.91 & 0.63 & 4.11 & 1.00$\times$ & -- & -- \\
 & Ours & 96.75 & 14.31 & 0.73 & 3.12 & 0.76$\times$ & -- & -- \\
\specialrule{0.3pt}{2pt}{1pt}
\multicolumn{9}{l}{\textit{DS-V4.1-Flash}} \\
\addlinespace[1pt]
\multirow{2}{*}{ReflAct} & Reactive & 78.20 & 31.34 & 3.30 & 11.14 & 1.00$\times$ & -- & -- \\
 & Ours & 78.76 & 21.11 & 0.81 & 6.09 & 0.55$\times$ & -- & -- \\
\addlinespace[1pt]
\multirow{2}{*}{MPO} & Reactive & 80.59 & 18.82 & 0.46 & 5.17 & 1.00$\times$ & -- & -- \\
 & Ours & 80.26 & 11.73 & 0.68 & 3.61 & 0.70$\times$ & -- & -- \\
\addlinespace[1pt]
\multirow{2}{*}{Reflexion} & Reactive & 87.08 & 84.12 & 2.03 & 23.06 & 1.00$\times$ & -- & -- \\
 & Ours & 85.39 & 61.16 & 2.35 & 17.64 & 0.76$\times$ & -- & -- \\
\specialrule{0.3pt}{2pt}{1pt}
\multicolumn{9}{l}{\textit{DeepSeek-V3.2}} \\
\addlinespace[1pt]
\multirow{2}{*}{ReflAct} & Reactive & 81.41 & 14.76 & 0.28 & 10.12 & 1.00$\times$ & -- & -- \\
 & Ours & 82.47 & 13.47 & 0.52 & 9.50 & 0.94$\times$ & -- & -- \\
\addlinespace[1pt]
\multirow{2}{*}{MPO} & Reactive & 86.12 & 16.22 & 0.21 & 11.02 & 1.00$\times$ & -- & -- \\
 & Ours & 87.19 & 11.33 & 0.44 & 7.99 & 0.73$\times$ & -- & -- \\
\addlinespace[1pt]
\multirow{2}{*}{Reflexion} & Reactive & 88.31 & 34.53 & 0.51 & 23.53 & 1.00$\times$ & -- & -- \\
 & Ours & 89.11 & 26.98 & 0.96 & 18.95 & 0.81$\times$ & -- & -- \\
\specialrule{0.8pt}{3pt}{2pt}
\multicolumn{9}{l}{\textbf{EB-Habitat}} \\
\multicolumn{9}{l}{\textit{Qwen3.6-Flash}} \\
\addlinespace[1pt]
\multirow{2}{*}{ReflAct} & Reactive & 58.49 & 29.03 & 1.29 & 6.13 & 1.00$\times$ & -- & -- \\
 & Ours & 59.10 & 16.48 & 1.81 & 4.56 & 0.74$\times$ & -- & -- \\
\addlinespace[1pt]
\multirow{2}{*}{Reflexion} & Reactive & 68.82 & 93.43 & 5.30 & 20.87 & 1.00$\times$ & -- & -- \\
 & Ours & 68.88 & 46.75 & 6.35 & 14.14 & 0.68$\times$ & -- & -- \\
\bottomrule
\end{tabular*}
\caption{Controller composition across three benchmarks with Reactive and EOCV (Ours) executors. Reward is on $[0,100]$; token totals are in millions (M). Qwen3.5 times are summed task durations in seconds. Ratios are relative to Reactive within each backbone and controller. Dashes denote unavailable values. DS-V4.1-Flash denotes DeepSeek-V4.1-Flash.}
\label{tab:controller_composition}
\label{tab:appendix_controller_token_usage}
\label{tab:ebhabitat_controller}
\end{table*}

\section{Additional Case Studies}
\label{sec:appendix_case_studies}

These additional case studies illustrate EOCV's execution dynamics and the sources of its performance and efficiency gains.

The ScienceWorld examples in Tables~\ref{tab:appendix_text_case} and~\ref{tab:appendix_text_case_dunk} use DeepSeek-V4.1-Flash and GPT-5.5, respectively.
In the first case, a cup cannot be placed in a closed oven, yet the naive batched executor continues with heating and inspection actions.
EOCV detects the mismatch, terminates the planned suffix, and opens the oven before retrying.
In the second case, both executors generate the same five-action plan to make sugar water.
When collecting water fails, the naive executor continues adding sugar and mixing an incomplete set of ingredients, whereas EOCV interrupts the batch and replans how to fill the cup.
These cases help explain how the EOCV-enhanced batched executor avoids cascading errors that can arise in naive batched execution.

The Qwen3.5-4B example in Table~\ref{tab:appendix_text_case_react} shows how the EOCV executor efficiently explores ALFWorld.
EOCV executes four consecutive information-seeking actions while skipping verification.
With new observations, the next executor call successfully picks up the watch.
EOCV executes the five actions shown from $t$ to $t{+}4$ in two executor calls, compared with five for ReAct.
This reduces weighted token usage from 2,554 to 1,432.

In the multimodal EB-Habitat benchmark, the environment returns image observations rather than textual feedback alone.
The Qwen3.6-Flash cases in Tables~\ref{tab:appendix_multimodal_case} (expanding Table~\ref{tab:case_study_multimodal}) and~\ref{tab:appendix_multimodal_case_efficiency} illustrate similar execution behavior with visual observations: EOCV verifies expected effects and replans when observations do not match expectations.

% Evidence: raw_results/text-case-studies.json, case closed-oven.
% Task: change-the-state-of-matter-of_27.
\begin{table*}[!t]
\centering
\fontsize{8}{9.5}\selectfont
\setlength{\tabcolsep}{3pt}
\renewcommand{\arraystretch}{1.12}
\begin{tabularx}{\textwidth}{@{}c Y Y Y Y Y@{}}
\toprule
 & \multicolumn{2}{c}{\textbf{Batched Executor}}
 & \multicolumn{3}{c}{\textbf{Ours}} \\
\cmidrule(lr){2-3} \cmidrule(lr){4-6}
\textbf{Step} & \textbf{Action} & \textbf{Observation}
              & \textbf{Action} & \textbf{Expectation} & \textbf{Observation} \\
\midrule
$t$ &
\texttt{move(cup, oven)} &
That can't be moved there, because the oven isn't open. &
\texttt{move(cup, oven)} &
You move the glass cup to the oven. &
That can't be moved there, because the oven isn't open. \\

$t{+}1$ &
\textcolor{mutedred}{\texttt{activate(oven)}} &
The oven is now activated. &
\textcolor{gray}{\texttt{close(oven)}} &
The oven is now closed. &
\textcolor{caseeocv}{\textbf{Terminate \& Replan}} \\

$t{+}2$ &
\textcolor{mutedred}{\texttt{wait()}} &
You decide to wait for 10 iterations. &
\textcolor{gray}{\texttt{activate(oven)}} &
The oven is now activated. &
---- \\

$t{+}3$ &
\textcolor{mutedred}{\texttt{look\_at(oven)}} &
An oven, which is turned on. The oven door is closed. &
\textcolor{gray}{\texttt{wait()}} &
You decide to wait for 10 iterations. &
---- \\
\bottomrule
\end{tabularx}
\caption{ScienceWorld case with DeepSeek-V4.1-Flash, expanding Table~\ref{tab:case_study_executor}. Slots preserve the order within each independently generated batch. \textcolor{casebaseline}{Red} actions continue after the failed move; \textcolor{casewithheld}{gray} actions are proposed but not executed. Dashes indicate no observation.\label{tab:appendix_text_case}}
\end{table*}

% Evidence: raw_results/text-case-studies.json, case sugar-water.
\begin{table*}[!t]
\centering
\fontsize{8}{9.5}\selectfont
\setlength{\tabcolsep}{3pt}
\renewcommand{\arraystretch}{1.12}
\begin{tabularx}{\textwidth}{@{}c Y Y Y Y Y@{}}
\toprule
 & \multicolumn{2}{c}{\textbf{Batched Executor}}
 & \multicolumn{3}{c}{\textbf{Ours}} \\
\cmidrule(lr){2-3} \cmidrule(lr){4-6}
\textbf{Step} & \textbf{Action} & \textbf{Observation}
              & \textbf{Action} & \textbf{Expectation} & \textbf{Observation} \\
\midrule
$t$ &
\texttt{activate(sink)} &
The sink is now activated. &
\texttt{activate(sink)} &
The sink is now activated. &
The sink is now activated. \\

$t{+}1$ &
\texttt{dunk(glass cup, sink)} &
The sink does not contain any liquids to dunk into. &
\texttt{dunk(glass cup, sink)} &
You dunk the glass cup in the sink. &
The sink does not contain any liquids to dunk into. \\

$t{+}2$ &
\textcolor{casebaseline}{\texttt{move(sugar, glass cup)}} &
You move the sugar to the glass cup. &
\textcolor{casewithheld}{\texttt{move(sugar, glass cup)}} &
You move the sugar to the glass cup. &
\textcolor{caseeocv}{\textbf{Terminate \& Replan}} \\

$t{+}3$ &
\textcolor{casebaseline}{\texttt{mix(glass cup)}} &
There is only one thing (sugar) &
\textcolor{casewithheld}{\texttt{mix(glass cup)}} &
Sugar and water mix to produce sugar water. &
\textemdash \\

$t{+}4$ &
\textcolor{casebaseline}{\texttt{focus\_on(sugar water)}} &
No known action matches that input. &
\textcolor{casewithheld}{\texttt{focus\_on(sugar water)}} &
You focus on the sugar water. &
\textemdash \\
\bottomrule
\end{tabularx}
\caption{ScienceWorld case with GPT-5.5. Slots preserve the order within each independently
generated batch. Red actions continue after the failed move; gray actions are proposed but not executed. Dashes
indicate no observation.\label{tab:appendix_text_case_dunk}}
\end{table*}

% Evidence: raw_results/text-case-studies.json, case watch-skip-replan.
% Token totals cover five complete ReAct calls and both complete EOCV calls,
% including EOCV actions omitted from the table or withheld from execution.
\begin{table*}[!t]
\centering
\fontsize{8}{9.5}\selectfont
\setlength{\tabcolsep}{3pt}
\renewcommand{\arraystretch}{1.12}
\begin{tabularx}{\textwidth}{@{}c Y Y Y Y Y@{}}
\toprule
 & \multicolumn{2}{c}{\shortstack{\textbf{ReAct}\\\textcolor{casebaseline}{5 executor calls; 2,554 weighted tokens}}}
 & \multicolumn{3}{c}{\shortstack{\textbf{Ours}\\\textcolor{caseeocv}{2 executor calls; 1,432 weighted tokens}}} \\
\cmidrule(lr){2-3} \cmidrule(lr){4-6}
\textbf{Step} & \textbf{Action} & \textbf{Observation}
              & \textbf{Action} & \textbf{Expectation} & \textbf{Observation} \\
\midrule
\multicolumn{3}{c}{} & \multicolumn{3}{c}{\textit{EOCV executor call 1}} \\
$t$ &
\texttt{goto(drawer 4)} &
You arrive at drawer 4. The drawer 4 is closed. &
\texttt{goto(drawer 4)} &
\textcolor{caseeocv}{\textit{Skip verification}} &
You arrive at drawer 4. The drawer 4 is closed. \\

$t{+}1$ &
\texttt{open(drawer 4)} &
You open the drawer 4. The drawer 4 is open. In it, you see nothing. &
\texttt{open(drawer 4)} &
\textcolor{caseeocv}{\textit{Skip verification}} &
You open the drawer 4. The drawer 4 is open. In it, you see nothing. \\

$t{+}2$ &
\texttt{goto(drawer 5)} &
You arrive at drawer 5. The drawer 5 is closed. &
\texttt{goto(drawer 5)} &
\textcolor{caseeocv}{\textit{Skip verification}} &
You arrive at drawer 5. The drawer 5 is closed. \\

$t{+}3$ &
\texttt{open(drawer 5)} &
You open the drawer 5. The drawer 5 is open. In it, you see a watch 1. &
\texttt{open(drawer 5)} &
\textcolor{caseeocv}{\textit{Skip verification}} &
You open the drawer 5. The drawer 5 is open. In it, you see a watch 1. \\

% Temporarily hidden; retain this unexecuted action for later restoration.
% --- & --- & --- &
% \textcolor{casewithheld}{\texttt{take(watch 1, drawer 1)}} &
% \textless unpredictable observation\textgreater &
% \textcolor{caseeocv}{\textbf{Misaligned: not executed}} \\
\midrule
\multicolumn{3}{c}{} & \multicolumn{3}{c}{\textit{EOCV executor call 2: first action shown}} \\
$t{+}4$ &
\texttt{take(watch 1, drawer 5)} &
You pick up the watch 1 from the drawer 5. &
\texttt{take(watch 1, drawer 5)} &
You pick up the watch 1 from the drawer 5. &
You pick up the watch 1 from the drawer 5. \\
\bottomrule
\end{tabularx}
\caption{ALFWorld case with Qwen3.5-4B. Token totals cover five complete ReAct calls and two complete EOCV calls, including EOCV actions omitted from the table.\label{tab:appendix_text_case_react}}
\end{table*}

\begin{table*}[!t]
\centering
\fontsize{8}{9.5}\selectfont
\setlength{\tabcolsep}{3pt}
\renewcommand{\arraystretch}{1.12}

\begin{tabularx}{\textwidth}{@{}c Y
>{\centering\arraybackslash}m{0.095\textwidth}
>{\raggedright\arraybackslash}m{0.12\textwidth}
Y >{\raggedright\arraybackslash}p{0.16\textwidth}
>{\centering\arraybackslash}m{0.095\textwidth}
>{\raggedright\arraybackslash}m{0.12\textwidth}@{}}
\toprule
 & \multicolumn{3}{c}{\textbf{Batched Executor}}
 & \multicolumn{4}{c}{\textbf{Ours}} \\
\cmidrule(lr){2-4} \cmidrule(lr){5-8}

\textbf{Step} & \textbf{Action} & \textbf{Observation} & \textbf{Annotation}
              & \textbf{Action} & \textbf{Expectation} & \textbf{Observation} & \textbf{Annotation} \\
\midrule

$t$ &
pick up the pear &
\includegraphics[width=0.085\textwidth]{figures/eb-habitat-counter.pdf} &
nothing is picked up &
pick up the pear &
The robot picks up a pear from the left counter. &
\includegraphics[width=0.085\textwidth]{figures/eb-habitat-counter.pdf} &
nothing is picked up \\

$t{+}1$ &
\textcolor{casebaseline}{navigate to the sofa} &
\includegraphics[width=0.085\textwidth]{figures/eb-habitat-sofa.pdf} &
agent moves to the sofa &
\textcolor{casewithheld}{navigate to the sofa} &
\textcolor{caseeocv}{\textit{Skip verification}} &
\textcolor{caseeocv}{\textbf{Terminate \& Replan}} &
\textemdash \\

$t{+}2$ &
\textcolor{casebaseline}{place at the sofa} &
\includegraphics[width=0.085\textwidth]{figures/eb-habitat-sofa.pdf} &
nothing can be placed &
\textcolor{casewithheld}{place at the sofa} &
The pear is placed on the sofa. &
\textemdash &
\textemdash \\
\bottomrule
\end{tabularx}
\caption{EB-Habitat case with Qwen3.6-Flash, expanding Table~\ref{tab:case_study_multimodal}. The batched executor continues executing the \textcolor{casebaseline}{red} actions after the visual observation contradicts the plan. EOCV detects the mismatch and withholds the \textcolor{casewithheld}{gray} actions. Annotations describe the images for the reader and are not available to the agent. Dashes indicate that an observation or annotation is not applicable.\label{tab:appendix_multimodal_case}}
% Because the pear is absent at the counter, the multi-step executor continues along an invalid trajectory, whereas EOCV detects the mismatch from the image observation and triggers replanning. Annotation denotes supplementary descriptions of visual observations for improved human readability and is not accessible to the agent; dashes indicate not applicable.
\end{table*}

% Image provenance: raw_results/eb-habitat-case-studies.json, plate-sink-to-right-counter.
\begin{table*}[!t]
\centering
\fontsize{8}{9.5}\selectfont
\setlength{\tabcolsep}{3pt}
\renewcommand{\arraystretch}{1.12}

\begin{tabularx}{\textwidth}{@{}c Y
>{\centering\arraybackslash}m{0.095\textwidth}
>{\raggedright\arraybackslash}m{0.12\textwidth}
Y >{\raggedright\arraybackslash}p{0.16\textwidth}
>{\centering\arraybackslash}m{0.095\textwidth}
>{\raggedright\arraybackslash}m{0.12\textwidth}@{}}
\toprule
 & \multicolumn{3}{c}{\shortstack{\textbf{ReAct}\\\textcolor{casebaseline}{3 executor calls; 2,476 weighted tokens}}}
 & \multicolumn{4}{c}{\shortstack{\textbf{Ours}\\\textcolor{caseeocv}{1 executor call; 968 weighted tokens}}} \\
\cmidrule(lr){2-4} \cmidrule(lr){5-8}

\textbf{Step} & \textbf{Action} & \textbf{Observation} & \textbf{Annotation}
              & \textbf{Action} & \textbf{Expectation} & \textbf{Observation} & \textbf{Annotation} \\
\midrule

$t$ &
place at table 2 &
\includegraphics[width=0.085\textwidth]{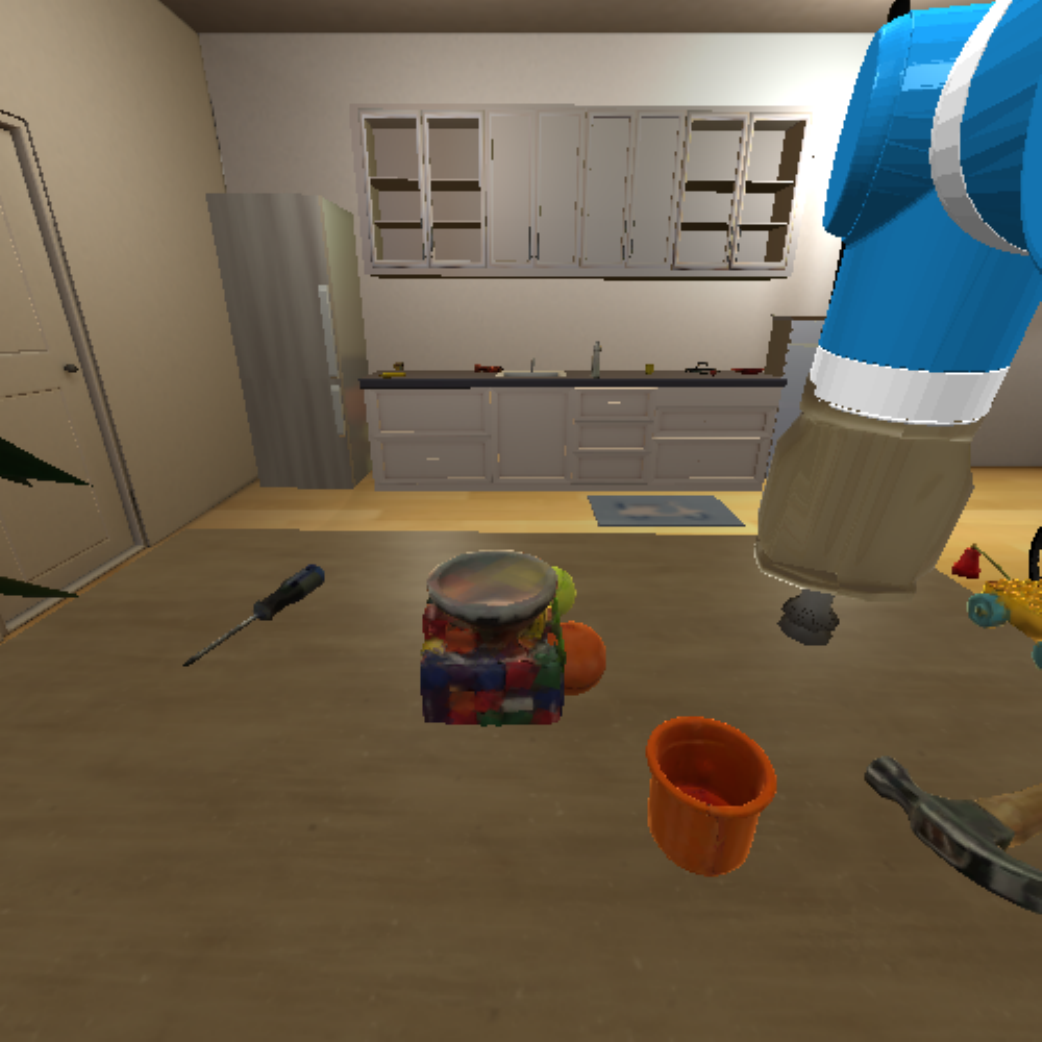} &
block is placed &
place at table 2 &
The block is placed on table 2. &
\includegraphics[width=0.085\textwidth]{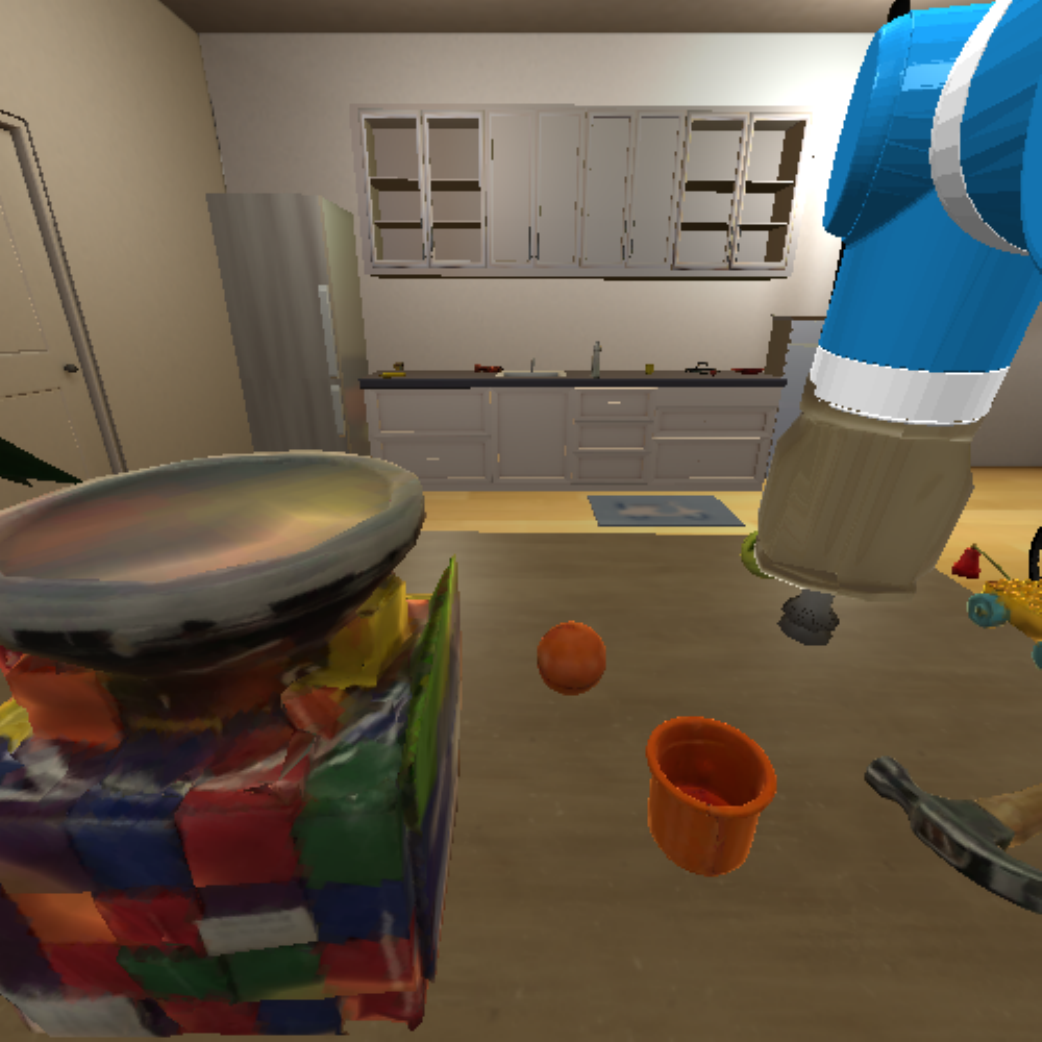} &
block is placed \\

$t{+}1$ &
pick up the hammer &
\includegraphics[width=0.085\textwidth]{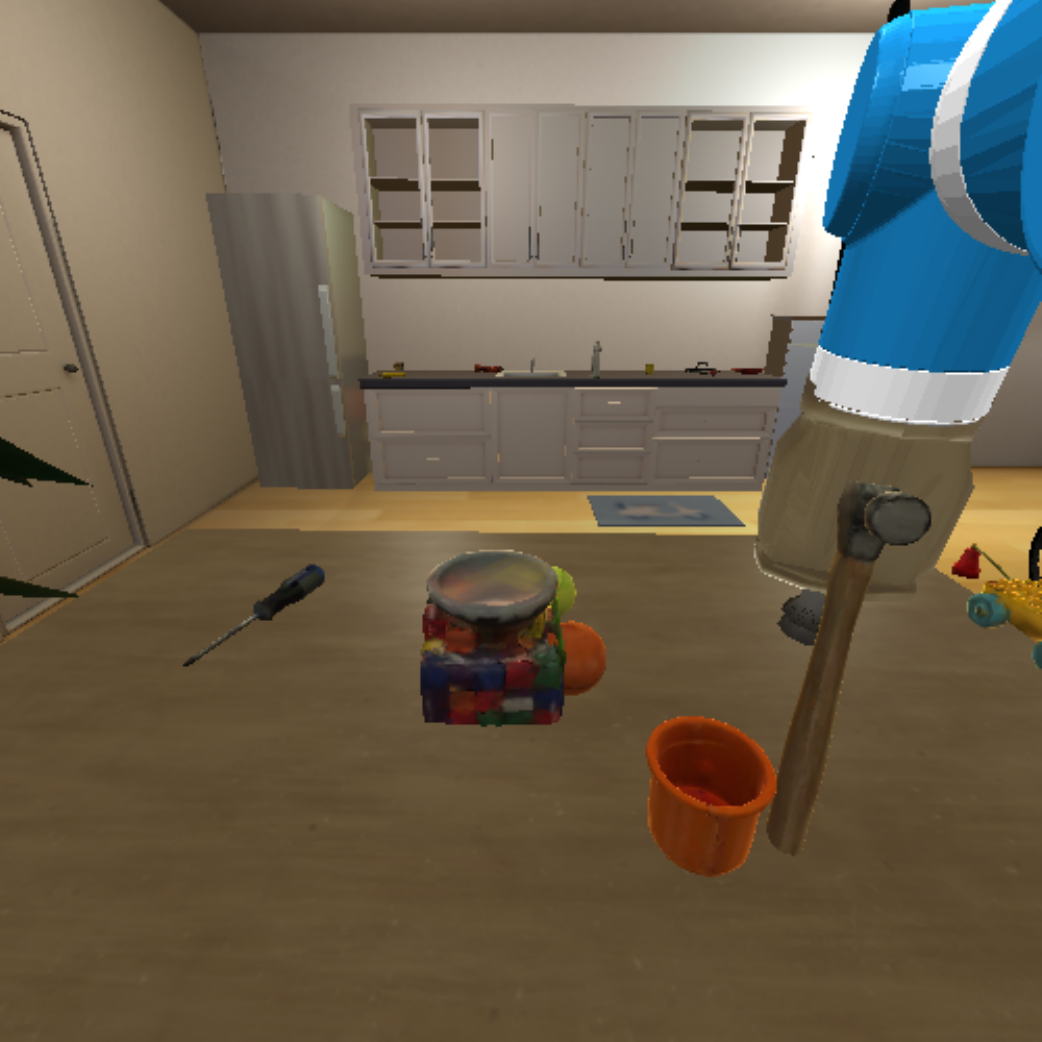} &
hammer is picked up &
pick up the hammer &
The robot is holding the hammer. &
\includegraphics[width=0.085\textwidth]{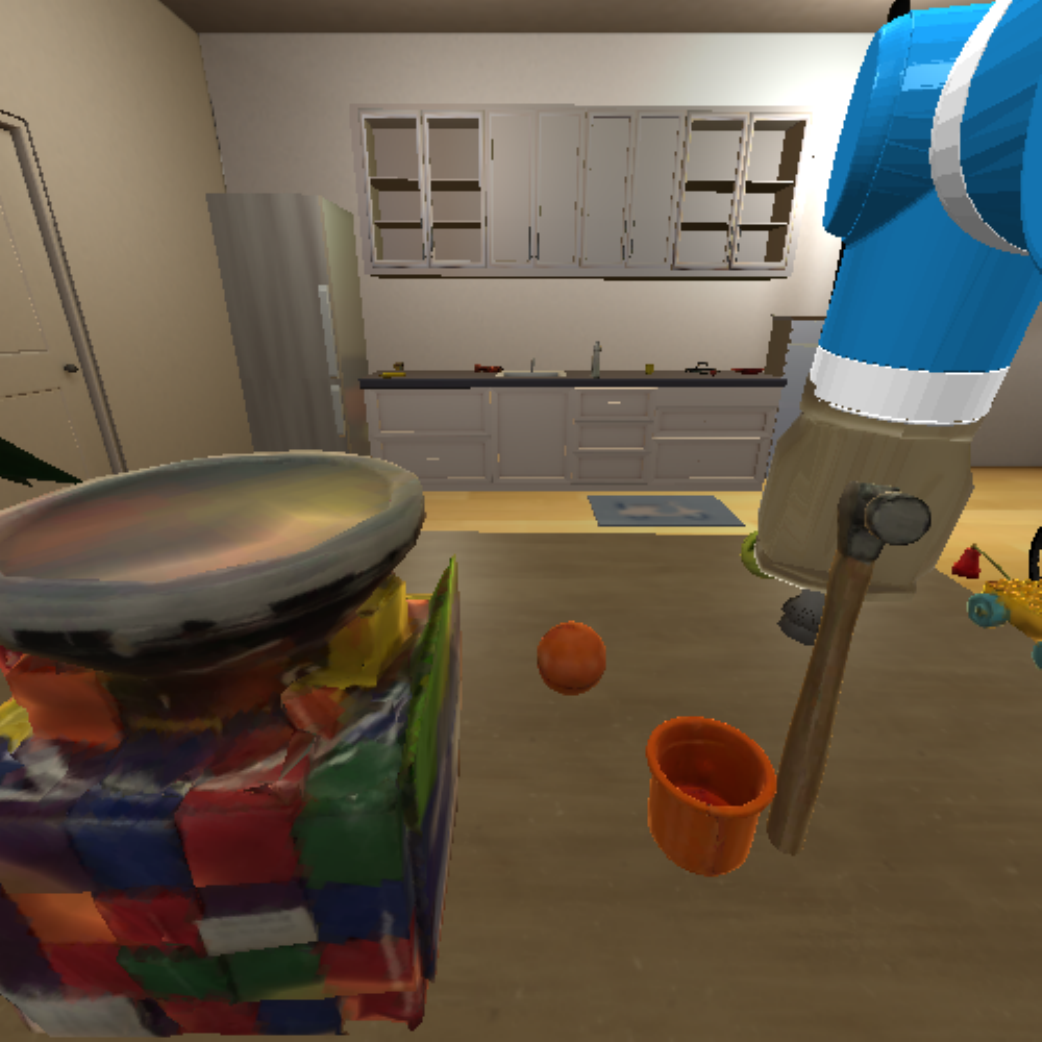} &
hammer is picked up \\

$t{+}2$ &
navigate to table 1 &
\includegraphics[width=0.085\textwidth]{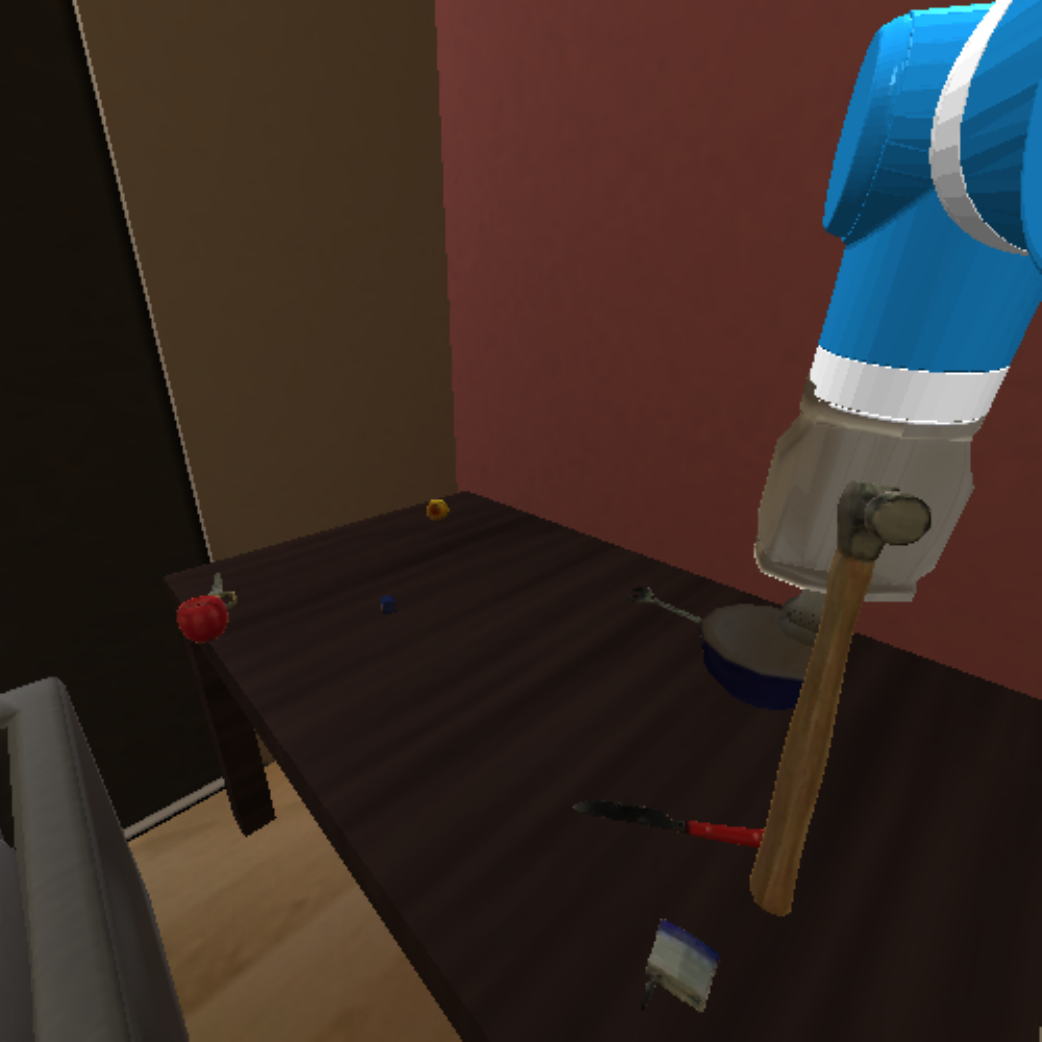} &
agent reaches table 1 &
navigate to table 1 &
\textcolor{caseeocv}{\textit{Skip verification}} &
\includegraphics[width=0.085\textwidth]{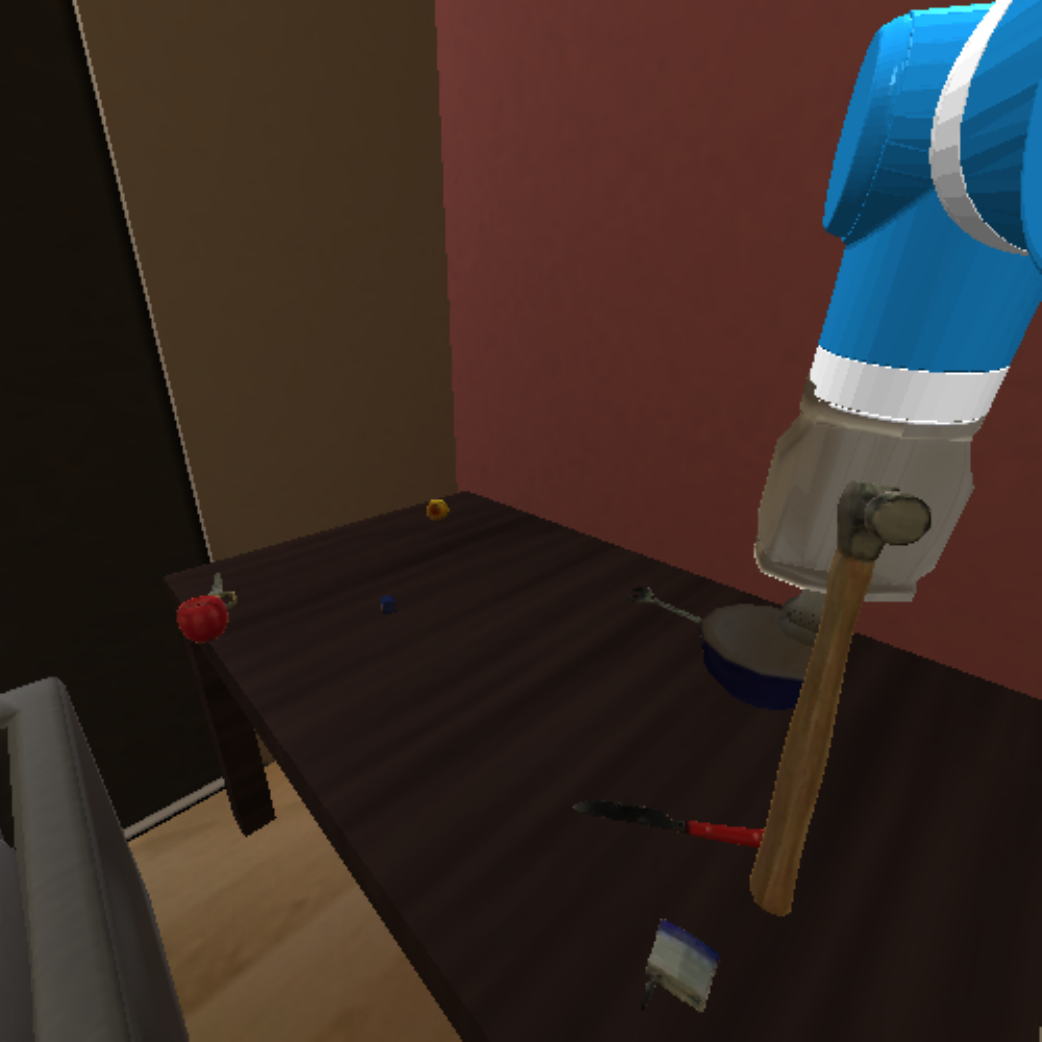} &
agent reaches table 1 \\
\bottomrule
\end{tabularx}
\caption{EB-Habitat efficiency case study with Qwen3.6-Flash. Token totals cover three complete ReAct calls and one complete EOCV calls. \label{tab:appendix_multimodal_case_efficiency}}
\end{table*}

% Keep the short implementation section together below the final case tables.
\appendixpagebreak
\appendixpagebreak
\section{AdaPlanner Implementation Details}
\label{sec:appendix_adaplanner}

We adapt the authors' ALFWorld implementation of AdaPlanner~\cite{sun2023adaplanner} to modern LLMs.
Complete prompts and output contracts appear in Appendix~\ref{sec:prompt_adaplanner}.

\paragraph{Prompts and decoding.}
We migrate the completion-based prompts to chat templates.
For a fair comparison, AdaPlanner uses the same in-context learning (ICL) example as the other executors, rewritten in AdaPlanner's format.

\paragraph{Evaluation settings.}
We evaluate the same 134 ALFWorld tasks on Qwen3.5-9B with reasoning enabled.
Each task permits at most 6 plan executions and 50 cumulative environment actions.

\paragraph{Token accounting.}
Totals include all model calls, including syntax checks, refinement, resume selection, and budget failures.
Reasoning is counted once as part of output tokens.

% Flush the notes and pending tables before restoring the document layout.
\clearpage
\endgroup
\onecolumn
\section{Prompts}
\label{sec:appendix_prompt}

This appendix presents the prompts used in our experiments for reproducibility.

\subsection{ALFWorld}

\subsubsection{Instructions}

\begin{promptbox}{Instructions for ALFWorld EOCV, Reactive, Batched}
Interact with a household to solve a task. Imagine you are an intelligent agent in a household environment and your target is to perform actions to complete the task goal.
At the beginning of your interactions, you will be given the detailed description of the current environment and your goal to accomplish.

The interaction is turn-based. Each turn is given in the following format:

```python
# Thought: your thoughts about the current condition and plan for your future actions to complete the current subgoal.
<your action>
# Observation: the feedback of your action
```

The available actions are:

inventory()
    Description: check your current inventory.
    Example:
        inventory() -> "You are carrying: a cup 4."


goto(receptacle)
    Description: move to a receptacle and observe what's in/on it.
    Parameters:
        receptacle (str): the receptacle to go to.
    Example:
        goto("sinkbasin 1") -> "You arrive at sinkbasin 1, on the sinkbasin 1, you see <receptacles>."


open(receptacle)
    Description: open a receptacle.
    Parameters:
        receptacle (str): the receptacle to open.
    Example:
        open("drawer 2") -> "You open the drawer 2. The drawer2 is open. In it, you see a creditcard 1, a dishsponge 1, and a spatula 2."


close(receptacle)
    Description: close a receptacle.
    Parameters:
        receptacle (str): the receptacle to close.
    Example:
        close("drawer 2") -> "You close the drawer 2."


take(object, receptacle)
    Description: take an object from a nearby receptacle.
    Parameters:
        object (str): the object to take.
        receptacle (str): the receptacle to take the object from.
    Example:
        take("spatula 1", "drawer 2") -> "You pick up the spatula 1 from the drawer 2."
    Hint:
        You can only take one thing with you at once.
        You must be near the receptacle to take something from it. That is, you should first "goto('receptacle A')" before "take(object, 'receptacle A')".
        If the receptacle is closed, you need to open it first.


move(object, receptacle)
    Description: place an object in or on a receptacle.
    Parameters:
        object (str): the object to move.
        receptacle (str): the receptacle to move the object to.
    Example:
        move("spatula 1", "drawer 2") -> "You move the spatula 1 to the drawer 2."
    Hint:
        You must be near the receptacle to move something to it.
        If the receptacle is closed, you need to open it first.


turn_on(object)
    Description: turn on an object.
    Parameters:
        object (str): the object to turn on.
    Example:
        turn_on('desklamp 1') -> "You turn on the desklamp 1."


heat(object, receptacle)
    Description: heat an object using a receptacle.
    Parameters:
        object (str): the object to heat.
        receptacle (str): the receptacle to use for heating.
    Example:
        heat("tomato 1", "microwave 1") -> "You heat the tomato 1 using the microwave 1."


clean(object, receptacle)
    Description: clean an object using a receptacle.
    Parameters:
        object (str): the object to clean.
        receptacle (str): the receptacle to use for cleaning.
    Example:
        clean("cup 1", "sinkbasin 1") -> "You clean the cup 1 using the sinkbasin 1."


cool(object, receptacle)
    Description: cool an object using a receptacle.
    Parameters:
        object (str): the object to cool.
        receptacle (str): the receptacle to use for cooling.
    Example:
        cool("cup 1", "fridge 1") -> "You cool the cup 1 using the fridge 1."


slice(object, tool)
    Description: slice an object using a sharp object.
    Parameters:
        object (str): the object to slice.
        tool (str): the sharp object to use for slicing.
    Example:
        slice("apple 1", "knife 1") -> "You slice the apple 1 using the knife 1."



Reminders:
1. There are several locations, each location can contain several objects. You can only take one object with you at once.
2. Actions are used to interact with the objects. The actions must be chosen from the given available actions. Any actions except provided available actions will be regarded as illegal. Furthermore, you can only interact with objects in the current location or the object you take.
3. The task uses strict matching. This means you must find exactly what the task requires, but not use a similar thing to replace it. For instance, if the task want a "knife", you cannot use a "butterknife" to replace it. Instead, you must find the object whose name is exactly "knife".
4. You can only take and keep one object in your inventory, so you have to put down the object you are holding before taking another one.

=========EOCV=========

Now, given the interaction history, you should give out a future action sequence towards the task goal.

Your response must strictly follow this format:

```python
# Thought: your thoughts about the current condition and plan for your future actions to complete the current subgoal.
<your action 1>
# Observation: <expected observation of your action>

# Thought: your thoughts about the current condition and plan for your future actions to complete the current subgoal.
<your action 2>
# Observation: <expected observation of your action>

# Thought: your thoughts about the current condition and plan for your future actions to complete the current subgoal.
<your action 3>
# Observation: <expected observation of your action>

<... repeat several times ...>

```

In order to verify if your purpose matches the actual environment observation, you are required to provide the expected observation of each action you take in the format above. For actions whose observations are unpredictable(e.g. "look", "examine", "goto", "open"). Simply write "<unpredictable observation>" as their expected observation.

In order to reduce cost, you are encouraged to generate a sequence of future actions. However, generating future actions without intermediate observations can make you deviate from the actual environment state. Therefore, stop extending the sequence when the environment state becomes unpredictable.
======================

=======Reactive=======
Now, given the interaction history, you should give out your next action towards the task goal.

Your response must strictly follow this format:

```python
# Thought: your thoughts about the current condition and plan for your future actions to complete the current subgoal.
<your action>
```
======================

=======Batched========
Now, given the interaction history, you should give out a future action sequence towards the task goal.

Your response must strictly follow this format:

```python
# Thought: your thoughts about the current condition and plan for your future actions to complete the current subgoal.
<your action 1>

# Thought: your thoughts about the current condition and plan for your future actions to complete the current subgoal.
<your action 2>

# Thought: your thoughts about the current condition and plan for your future actions to complete the current subgoal.
<your action 3>

<... repeat several times ...>

```
======================
Here is an example:

{icl_example}

=======Reactive=======
Now, here is the real task, think carefully about the task, then give out your next action(give only one action) in the same format as the example above.
======================

=====EOCV/Batched=====
Now, here is the real task, think carefully about the task, then give out a future action sequence in the same format as the example above.

You are required to plan at most {max_planning_steps} future actions per generation step in this task. If you are not sure about the environment state, stop extending the action sequence.
======================
{interact_history}

\end{promptbox}

\subsubsection{ICL Examples}

\begin{promptbox}{ICL Example for ALFWorld EOCV}
```python
# Task Description: -= Welcome to TextWorld, ALFRED! =-

You are in the middle of a room. Looking quickly around you, you see a cabinet 11, a cabinet 10, a cabinet 9, a cabinet 8, a cabinet 7, a cabinet 6, a cabinet 5, a cabinet 4, a cabinet 3, a cabinet 2, a cabinet 1, a coffeemachine 1, a countertop 2, a countertop 1, a diningtable 1, a drawer 3, a drawer 2, a drawer 1, a fridge 1, a garbagecan 1, a microwave 1, a sinkbasin 1, a stoveburner 4, a stoveburner 3, a stoveburner 2, a stoveburner 1, and a toaster 1.\n\nYour task is to: put a hot bread in garbagecan.

# Thought: I need to find bread, which is likely to be found in a fridge.
goto("fridge 1")
# Observation: You arrive at fridge 1. The fridge 1 is closed.

# Thought: I should open the fridge to see if there is any bread inside.
open("fridge 1")
# Observation: You open the fridge 1. The fridge 1 is open. In it, you see a cup 2, a egg 3, a egg 2, a egg 1, a lettuce 3, a mug 3, and a tomato 2.

# Thought: There is no bread in the fridge. I should look around the countertops and see if I can find bread.
goto("countertop 1")
# Observation: You arrive at countertop 1. On the countertop 1, you see a bread 1, a dishsponge 1, a knife 1, a mug 1, a peppershaker 1, a potato 1, and a spoon 1.

# Thought: I see bread 1 on countertop 1. I will pick up bread 1 and heat it using the microwave.
take("bread 1", "countertop 1")
# Observation: You pick up the bread 1 from the countertop 1.

# Thought: I've picked up bread 1. Now I need to go to the microwave to heat it.
goto("microwave 1")
# Observation: You arrive at microwave 1. The microwave 1 is closed.

# Thought: I need to open the microwave first before heating the bread.
open("microwave 1")
# Observation: You open the microwave 1. The microwave 1 is open. In it, you see nothing.
```
Your response can be:

```python

# Thought: Now I should heat the bread using the microwave.
heat("bread 1", "microwave 1")
# Observation: You heat the bread 1 using the microwave 1.

# Thought: The bread is now hot. I need to put the hot bread in the garbagecan to complete the task.
goto("garbagecan 1")
# Observation: <unpredictable observation>

```
\end{promptbox}
\begin{promptbox}{ICL Example for ALFWorld Reactive}
```python
# Task Description: -= Welcome to TextWorld, ALFRED! =-

You are in the middle of a room. Looking quickly around you, you see a cabinet 11, a cabinet 10, a cabinet 9, a cabinet 8, a cabinet 7, a cabinet 6, a cabinet 5, a cabinet 4, a cabinet 3, a cabinet 2, a cabinet 1, a coffeemachine 1, a countertop 2, a countertop 1, a diningtable 1, a drawer 3, a drawer 2, a drawer 1, a fridge 1, a garbagecan 1, a microwave 1, a sinkbasin 1, a stoveburner 4, a stoveburner 3, a stoveburner 2, a stoveburner 1, and a toaster 1.\n\nYour task is to: put a hot bread in garbagecan.

# Thought: I need to find bread, which is likely to be found in a fridge.
goto("fridge 1")
# Observation: You arrive at fridge 1. The fridge 1 is closed.

# Thought: I should open the fridge to see if there is any bread inside.
open("fridge 1")
# Observation: You open the fridge 1. The fridge 1 is open. In it, you see a cup 2, a egg 3, a egg 2, a egg 1, a lettuce 3, a mug 3, and a tomato 2.

# Thought: There is no bread in the fridge. I should look around the countertops and see if I can find bread.
goto("countertop 1")
# Observation: You arrive at countertop 1. On the countertop 1, you see a bread 1, a dishsponge 1, a knife 1, a mug 1, a peppershaker 1, a potato 1, and a spoon 1.

# Thought: I see bread 1 on countertop 1. I will pick up bread 1 and heat it using the microwave.
take("bread 1", "countertop 1")
# Observation: You pick up the bread 1 from the countertop 1.

# Thought: I've picked up bread 1. Now I need to go to the microwave to heat it.
goto("microwave 1")
# Observation: You arrive at microwave 1. The microwave 1 is closed.

# Thought: I need to open the microwave first before heating the bread.
open("microwave 1")
# Observation: You open the microwave 1. The microwave 1 is open. In it, you see nothing.

# Thought: Now I should heat the bread using the microwave.
heat("bread 1", "microwave 1")
# Observation: You heat the bread 1 using the microwave 1.
```
Your response can be:

```python
# Thought: The bread is now hot. I need to put the hot bread in the garbagecan to complete the task.
goto("garbagecan 1")
```
\end{promptbox}
\begin{promptbox}{ICL Example for ALFWorld Batched}
```python
# Task Description: -= Welcome to TextWorld, ALFRED! =-

You are in the middle of a room. Looking quickly around you, you see a cabinet 11, a cabinet 10, a cabinet 9, a cabinet 8, a cabinet 7, a cabinet 6, a cabinet 5, a cabinet 4, a cabinet 3, a cabinet 2, a cabinet 1, a coffeemachine 1, a countertop 2, a countertop 1, a diningtable 1, a drawer 3, a drawer 2, a drawer 1, a fridge 1, a garbagecan 1, a microwave 1, a sinkbasin 1, a stoveburner 4, a stoveburner 3, a stoveburner 2, a stoveburner 1, and a toaster 1.\n\nYour task is to: put a hot bread in garbagecan.

# Thought: I need to find bread, which is likely to be found in a fridge.
goto("fridge 1")
# Observation: You arrive at fridge 1. The fridge 1 is closed.

# Thought: I should open the fridge to see if there is any bread inside.
open("fridge 1")
# Observation: You open the fridge 1. The fridge 1 is open. In it, you see a cup 2, a egg 3, a egg 2, a egg 1, a lettuce 3, a mug 3, and a tomato 2.

# Thought: There is no bread in the fridge. I should look around the countertops and see if I can find bread.
goto("countertop 1")
# Observation: You arrive at countertop 1. On the countertop 1, you see a bread 1, a dishsponge 1, a knife 1, a mug 1, a peppershaker 1, a potato 1, and a spoon 1.

# Thought: I see bread 1 on countertop 1. I will pick up bread 1 and heat it using the microwave.
take("bread 1", "countertop 1")
# Observation: You pick up the bread 1 from the countertop 1.

# Thought: I've picked up bread 1. Now I need to go to the microwave to heat it.
goto("microwave 1")
# Observation: You arrive at microwave 1. The microwave 1 is closed.

# Thought: I need to open the microwave first before heating the bread.
open("microwave 1")
# Observation: You open the microwave 1. The microwave 1 is open. In it, you see nothing.
```
Your response can be:

```python

# Thought: Now I should heat the bread using the microwave.
heat("bread 1", "microwave 1")

# Thought: The bread is now hot. I need to put the hot bread in the garbagecan to complete the task.
goto("garbagecan 1")
```
\end{promptbox}

% Generated by scripts/export_adaplanner_prompts.py; do not edit prompt text here.
\subsection{AdaPlanner on ALFWorld}
\label{sec:prompt_adaplanner}

These are the prompts for the AdaPlanner implementation in Appendix~\ref{sec:appendix_adaplanner}.

\subsubsection{Instruction and ICL example}

The Instruction serves as system prompt, and are used by 
the following syntax checking, plan refinement, resume selection and resume selection progress.

\begin{promptbox}{Instruction for Adaplanner}
You are a household agent. Here is some Python code defining a household environment:

Use literal_eval to convert the answer from ask() to a list.
literal_eval is provided by the runtime.

In the environment, you can ask questions to an assistant by ask():
for example: You have a list of receptacles, 
and you want to sort them by the likelihood of a soapbar appearing in them. 
You can do this by asking the assistant:
```
receptacles = ['countertop 1', 'garbagecan 1', 'sinkbasin 2', 'sinkbasin 1', 'toilet 1', 'toiletpaperhanger 1', 'towelholder 1']
answer = ask(f'Sort the list of receptacles, starting from the one a soapbar is most likely to appear: {receptacles}. You should return a Python list.')
# answer = ['sinkbasin 1', 'sinkbasin 2', 'countertop 1', 'towelholder 1', 'toiletpaperhanger 1', 'garbagecan 1', 'toilet 1']
```

Agent class represents the state of the agent, including its location,
what it's holding as well as the actions it can take.
```
class Agent:
    def __init__(self, receptacles):
        self.location = None
        self.holding = None
        self.receptacles = receptacles

    # Here are the admissible actions the agent can take:
    
    # Go to a receptacle and update the agent's location. 
    # For example, 'On the countertop 1, you see a candle 1, a cloth 2, and a soapbar 1.' = goto('countertop 1')
    # For example, 'On the sidetable 2, you see nothing.' = goto('sidetable 2')
    def goto(self, receptacle):
        ...

    # Take an object from a receptacle if the agent is not holding anything. 
    # For example, 'You pick up the soapbar 1 from the towelholder 1.' = take('soapbar 1', 'towelholder 1')
    def take(self, object, receptacle):
        ...
        
    # Put an object in or on a receptacle if the agent is holding it. 
    # For example, 'You put the soapbar 1 in/on the cabinet 1.' = put('soapbar 1', 'cabinet 1')
    def put(self, object, receptacle):
        ...

    # Open a receptacle and observe its contents. 
    # For example, 'You open the cabinet 1. The cabinet 1 is open. In it, you see a cloth 1.' = open_receptacle('cabinet 1')
    def open_receptacle(self, receptacle):
        ...

    # Clean an object with a receptacle. 
    # For example, 'You clean the soapbar 1 using the sinkbasin 1.' = clean('soapbar 1', 'sinkbasin 1')
    def clean(self, object, receptacle):
        ...

    # Heat an object with a receptacle. 
    # For example, 'You heat the tomato 1 using the microwave 1.' = heat('tomato 1', 'microwave 1')
    def heat(self, object, receptacle):
        ...

    # Cool an object with a receptacle. 
    # For example, 'You cool the pan 2 using the fridge 1.' = cool('pan 2', 'fridge 1')
    def cool(self, object, receptacle):
        ...

    # Turn on an object. 
    # For example, 'You turn on the desklamp 1.' = turn_on('desklamp 1')
    def turn_on(self, object):
        ...

    # Report agent's current state, including its location, what it's holding, and last action and observation.
    # This function should only be used in assertion.
    def report(self):
        ...
```
\end{promptbox}

\begin{promptbox}{ICL example and task description for Adaplanner}
Here is an example of a solution to the task:
```
# -= Welcome to TextWorld, ALFRED! =-
#
# You are in the middle of a room. Looking quickly around you, you see a cabinet 11, a cabinet 10, a cabinet 9, a cabinet 8, a cabinet 7, a cabinet 6, a cabinet 5, a cabinet 4, a cabinet 3, a cabinet 2, a cabinet 1, a coffeemachine 1, a countertop 2, a countertop 1, a diningtable 1, a drawer 3, a drawer 2, a drawer 1, a fridge 1, a garbagecan 1, a microwave 1, a sinkbasin 1, a stoveburner 4, a stoveburner 3, a stoveburner 2, a stoveburner 1, and a toaster 1.
#
# Your task is to: put a hot bread in garbagecan.
# Here is a solution:

def solution(agent, start_from=1):
    # General plan: find bread, take it, heat it in the microwave,
    # and put the hot bread in the garbagecan.
    if start_from <= 1:
        print("[Step 1] Find bread")
        # Bread may be in the fridge; if it is absent, check the countertop.
        for receptacle in ["fridge 1", "countertop 1"]:
            observation = agent.goto(receptacle)
            assert agent.location == receptacle, f"Error in [Step 1]: {agent.report()}"
            if "closed" in observation:
                observation = agent.open_receptacle(receptacle)
            if "bread " in observation:
                break
        assert "bread " in observation, f"Error in [Step 1]: Bread was not found. {agent.report()}"

    if start_from <= 2:
        print("[Step 2] Take the bread")
        # Use the observed identifier to pick up the bread that was found.
        answer = ask(f"From this observation, what is the identifier of the bread? {observation} Return only the number.")
        bread = f"bread {answer}"
        observation = agent.take(bread, receptacle)
        assert agent.holding == bread, f"Error in [Step 2]: {agent.report()}"

    if start_from <= 3:
        print("[Step 3] Heat the bread")
        # Go to the microwave and open it if closed, then heat the held bread.
        observation = agent.goto("microwave 1")
        assert agent.location == "microwave 1", f"Error in [Step 3]: {agent.report()}"
        if "closed" in observation:
            observation = agent.open_receptacle("microwave 1")
            assert "microwave 1 is open" in observation, f"Error in [Step 3]: {agent.report()}"
        observation = agent.heat(bread, "microwave 1")
        assert f"You heat the {bread} using the microwave 1." in observation, f"Error in [Step 3]: {agent.report()}"

    if start_from <= 4:
        print("[Step 4] Put the hot bread in the garbagecan")
        # The goal requires placing the heated bread in the garbagecan.
        observation = agent.goto("garbagecan 1")
        assert agent.location == "garbagecan 1", f"Error in [Step 4]: {agent.report()}"
        observation = agent.put(bread, "garbagecan 1")
        assert f"You put the {bread} in/on the garbagecan 1." in observation, f"Error in [Step 4]: {agent.report()}"
```
Here is the actual task:
receptacles = <RECEPTACLES>
<TASK_DESCRIPTION>
Write only the complete Python solution function for this task
\end{promptbox}

\subsubsection{Syntax Checking}

\begin{promptbox}{Syntax check prompt for Adaplanner}
Check the syntax of a Python solution function. 
Return {"decision":"No"} if there are no syntax errors. 
If there are syntax errors, return 
{"decision":"Yes","code":["def solution(agent, start_from=1):","    ..."]}. 
Each code element is exactly one Python source line;
preserve leading indentation spaces. Check Python syntax only. 
The runtime provides agent, ask, literal_eval and common Python builtins; 
missing imports or runtime names are not syntax errors. 
For No, omit code and do not copy the function. 
Keep the algorithm and assertions unchanged. No other prose.

Here is the python code: 
<SOLUTION_FUNCTION>

Question 1: Are there any syntax errors present in the code? Answer Yes/No.
Answer Question 1 as a JSON object with decision equal to "Yes" or "No". 

Question 2 applies ONLY if there are syntax errors: 
output the entire corrected function as an array of source lines in code, 
preserving indentation. 
Otherwise output only {"decision":"No"}.
\end{promptbox}

\subsubsection{Plan Refinement}

\begin{promptbox}{Plan refinement prompt for Adaplanner}
<PREVIOUS_SOLUTION>

You have generated code of solution() to solve the task. However, you executed the solution() function and get an error message:
<EXECUTION_ERROR>

[Current agent state]
<CURRENT_STATE>
[Saved local variables]
<SAVED_LOCALS>
[Actual interactions]
<ACTION_OBSERVATION_TRAJECTORY>

Referring to the successful case and the error message,
revise the plan and return only the complete corrected Python function, 
preserving newlines and indentation. 
The environment has NOT been reset. 
\end{promptbox}

\subsubsection{Resume Selection}

\begin{promptbox}{Resume selection for ALFWorld}
Previously, you generated some code defining a solution function as in [Previous solution]. 
The previous code is executed and outputs some error. 
Now you just revised the code as in [Revised solution]. 
Determine from which step these two version differs. 
Return only a JSON object with the positive integer field start_from.

Choose a safe numbered step for resuming a revised plan in the current environment. 
Return JSON {"start_from": N} with a positive integer N. 
Resume no later than the failed step, taking saved local variables into account. 
If unsure, return {"start_from":1}.

[Previous solution]
<PREVIOUS_SOLUTION>

[Revised solution]
<REVISED_SOLUTION>
[Execution error]
<EXECUTION_ERROR>
[Saved locals]
<SAVED_LOCALS>
\end{promptbox}

\subsubsection{In-Plan Reasoning}

Note that this progress does not adapt the general instruction as system prompt.

\begin{promptbox}{In-plan reasoning for AdaPlanner}
You are the household agent's in-plan reasoning assistant.
Answer the question using only the supplied observations and receptacles.
Return a JSON object with exactly one field, answer. 
Interpret output-format requests in the question as applying to the answer value; 
the runtime converts arrays to Python list literals and IDs to strings for ask(). 
No Markdown fences or additional explanation.

<IN_PLAN_QUESTION>
\end{promptbox}

\subsection{ScienceWorld}

\subsubsection{Instructions}

\begin{promptbox}{Instruction for ScienceWorld EOCV, Reactive, Batched}
Interact with a household to solve a task. Imagine you are an intelligent agent in a household environment and your target is to perform actions to complete the task goal.
At the beginning of your interactions, you will be given the detailed description of the current environment and your goal to accomplish.

The interaction is turn-based. Each turn is given in the following format:

```python
# Thought: your thoughts about the current condition and plan for your future actions to complete the current subgoal.
<your action>
# Observation: the feedback of your action
```

The available actions are:

activate(obj: str)
    Description: Activate a device.
    Parameters:
        obj (str): The name of the device to activate.
    Example:
        activate('sink') -> "The sink is now activated."


choose(number: int)
    Description: When an ambiguous request occurs, the environment feedback will be "This is an ambiguous request. Please use the 'choose(number)' action to select the
specific action you intended."(followed by multiple possible actions). Use "choose" to select the specific action, instead of directly outputting the number.It is prohibited
to use "choose" at other times.
    Example:
        choose(1) -> [Executes the first action from the ambiguous options]


close(obj: str)
    Description: Close a container.
    Parameters:
        obj (str): The name of the container to close.
    Example:
        close('door to hallway') -> "The door is now closed."


deactivate(obj: str)
    Description: Deactivate a device.
    Parameters:
        obj (str): The name of the device to deactivate.
    Example:
        deactivate('sink') -> "The sink is now deactivated."


dunk(obj: str, target: str)
    Description: transferring liquid from target to the obj.
    Parameters:
        obj (str): The empty container.
        target (str): Another container to dunk into(containing liquid).
    Example:
        dunk('glass cup', 'sink') -> "You dunk the glass cup in the sink."


focus_on(obj: str)
    Description: Signal intent on a task object. This action is used to evaluate your performance, it should be called only if the task description requires you to focus
something. Focus on wrong/non-existence object can cause the failure of current task, so it is strongly recommended to use "look_around" to see what's in the environment
before use "focus_on".
    Parameters:
        obj (str): The task-related object to focus on.
    Example:
        focus_on('red light bulb') -> "You focus on the red light bulb."


look_around()
    Description: Describe the current room.
    Example:
        look_around() -> "[description of surroundings including furniture and objects]"


look_at(obj: str)
    Description: Describe a container's contents.
    Special: "look at inventory" to check your inventory.
    Parameters:
        obj (str): The container to look inside.
    Example:
        look_at('cupboard') -> "The cupboard contains a plate and a bowl."


mix(obj: str)
    Description: Chemically mix a container(move the ingredients to the container before mixing).
    Parameters:
        obj (str): The container to mix.
    Example:
        mix('bowl') -> "Blue and yellow paint mix to produce green paint."


move(obj: str, target: str)
    Description: Move an object to a container. (Can't be used to move object to another room)
    Parameters:
        obj (str): The object to move.
        target (str): The destination container.
    Example:
        move('metal pot', 'sink') -> "You move the metal pot to the sink."


open(obj: str)
    Description: Open a container(You should use look_around later to check what's in the container).
    Parameters:
        obj (str): The name of the container to open.
    Example:
        open('cupboard') -> "The cupboard is now open."


pick_up(obj: str)
    Description: Move an object in the current room to the inventory.
    Parameters:
        obj (str): The object to pick up.
    Example:
        pick_up('thermometer') -> "You move the thermometer to the inventory."


pour(obj: str, target: str)
    Description: Pour the liquid in one container into another container.
    Parameters:
        obj (str): The container to pour(Should be moveable).
        target (str): Another container to pour into.
    Example:
        pour('jug', 'flower pot 3') -> "You pour the contents of the jug into the flower pot 3."


read(obj: str)
    Description: Read a note or book.
    Parameters:
        obj (str): The object to read.
    Example:
        read('recipe') -> "The recipe reads: ...")


teleport(loc: str)
    Description: Teleport to a specific room or outside.
    Parameters:
        loc (str): The name of the room to teleport to.
    Example:
        teleport('kitchen') -> "You teleport to the kitchen."


use(obj: str, target: str = None)
    Description: Use a tool(on the target if it exists). Note that this action is for high-level actions such as dig a hole or plant a seed. You should use other actions
first(such as move, activate, pour, etc.) unless there is no other available actions to complete the what you want to do.
    Parameters:
        obj (str): The tool to use.
        target (str, optional): The target object to use the tool on.
    Example:
        use('thermometer', 'glass cup') -> "the thermometer measures a temperature of 107 degrees celsius"


wait()
    Description: Do nothing for 10 steps.
    Example:
        wait() -> "You decide to wait for 10 iterations."

wait1()
    Description: Do nothing for 1 step.
    Example:
        wait1() -> "You decide to wait for 1 iterations."


Reminders:
1. The environment includes the following rooms: "art studio", "greenhouse", "workshop", "kitchen", "bedroom", "living room", "hallway", "foundry", and "outside". Use "teleport" to teleport to a specific room.
2. There will be several objects in each room to interact with. And you will have an inventory to keep some objects.
3. Actions are used to interact with the objects. The actions must be chosen from the given available actions. Any actions except provided available actions will be regarded as illegal. Furthermore, you can only interact with objects in the current room or your inventory.
4. The environment automatically shuts down when the task is completed, so whenever you are required to perform more actions, the task is not done.

Now, given the interaction history, you should give out a future action sequence towards the task goal.

Your response must strictly follow this format:
=========EOCV=========
```python
# Thought: your thoughts about the current condition and plan for your future actions to complete the current subgoal.
<your action 1>
# Observation: <expected observation of your action>

# Thought: your thoughts about the current condition and plan for your future actions to complete the current subgoal.
<your action 2>
# Observation: <expected observation of your action>

# Thought: your thoughts about the current condition and plan for your future actions to complete the current subgoal.
<your action 3>
# Observation: <expected observation of your action>

<... repeat several times ...>

```
In order to verify if your purpose matches the actual environment observation, you are required to provide the expected observation of each action you take in the format above. For actions whose observations are unpredictable(e.g. "look_around", "look_at", "choose", "read"), simply write "<unpredictable observation>" as their expected observations.

In order to reduce cost, you are encouraged to generate a sequence of future actions. However, generating future actions without intermediate observations can make you deviate from the actual environment state. Therefore, stop extending the sequence when the environment state becomes unpredictable.
======================

=======Reactive=======
```python
# Thought: your thoughts about the current condition and plan for your future actions to complete the current subgoal.
<your action>
```
======================

=======Batched========
Your response must strictly follow this format:

```python
# Thought: your thoughts about the current condition and plan for your future actions to complete the current subgoal.
<your action 1>

# Thought: your thoughts about the current condition and plan for your future actions to complete the current subgoal.
<your action 2>

# Thought: your thoughts about the current condition and plan for your future actions to complete the current subgoal.
<your action 3>

<... repeat several times ...>
```
======================

Here is an example:

<icl_example>

=======Reactive=======
Now, here is the real task, think carefully about the task, then give out your next action(give only one action) in the same format as the example above. Take experiment or use your common sense to solve the task are all acceptable. Keep in mind that you only have 70 step action budget to complete the task.
======================

=====EOCV/Batched=====
Now, here is the real task, think carefully about the task, then give out a future action sequence in the same format as the example above. Take experiment or use your common sense to solve the task are all acceptable. Keep in mind that you only have 70 step action budget to complete the task.


You should plan no more than {max_planning_steps} future actions per generation step in this task, but you do not have to generate that many every time. If you are not sure about the environment state, stop extending the action sequence.
======================

{interact_history}
\end{promptbox}

\subsubsection{ICL Examples}

\begin{promptbox}{ICL Example for ScienceWorld EOCV}
```python
# Task Description: Your task is to use chemistry to create green paint. When you are done, focus on the green paint.

# Thought: I need to find materials to create green paint, which are likely to be found in an art studio.
look_around()
# Observation: This room is called the bedroom. In it, you see:
    the agent
    a substance called air
    a bed. On the bed is: a mattress. On the mattress is: a white pillow..
    a closet. The closet door is open. In the closet is: nothing.
    a picture
    a table. On the table is: nothing.

# Thought: The bedroom doesn't have the materials I need. I should go to a place where paint and painting supplies are likely to be found, such as an art studio.
teleport("art studio")
# Observation: You teleport to the art studio.

# Thought: Now that I'm in the art studio, I should look around to find the necessary materials to create green paint.
look_around()
# Observation: This room is called the art studio. In it, you see:
    the agent
    a substance called air
    a large cupboard. The large cupboard door is open. In the large cupboard is: a drawer, a wood cup (containing yellow paint), a wood cup (containing blue paint), a wood cup (containing red paint).
    a table. On the table is: a glass cup (containing nothing).
    a wood cup (containing yellow paint)
    a wood cup (containing blue paint)
    a wood cup (containing red paint)
```

Your response can be:

```python
# Thought: I see blue paint here, which is one of the primary colors needed to mix green paint. I should start by pouring it into an empty container to mix with another color.
pour("cup containing blue paint", "glass cup")
# Observation: You pour the contents of the wood cup into the glass cup.

# Thought: Yellow paint is the other primary color needed to mix with blue to create green. I should add the yellow paint to the container with blue paint.
pour("cup containing yellow paint", "glass cup")
# Observation: You pour the contents of the wood cup into the glass cup.

# Thought: I need to mix the blue and yellow paint to create green paint.
mix("glass cup")
# Observation: Blue and yellow paint mix to produce green paint.

# Thought: The task is to create green paint and then focus on it. Now that the green paint is made. Before I focus on the green paint, I should look at the glass cup to confirm it contains green paint.
look_at("glass cup")
# Observation: <unpredictable observation>
```

If the `look_at("glass cup")` observes that "The glass cup contains green paint.", then your next response can be:
```python
# Thought: The task is to create green paint and then focus on it. Now that the green paint is made, I should concentrate on the final product.
focus_on("green paint")
# Observation: You focus on the green paint.
```
\end{promptbox}
\begin{promptbox}{ICL Example for ScienceWorld Reactive}
```python
# Task Description: Your task is to use chemistry to create green paint. When you are done, focus on the green paint.

# Thought: I need to find materials to create green paint, which are likely to be found in an art studio.
look_around()
# Observation: This room is called the bedroom. In it, you see:
    the agent
    a substance called air
    a bed. On the bed is: a mattress. On the mattress is: a white pillow..
    a closet. The closet door is open. In the closet is: nothing.
    a picture
    a table. On the table is: nothing.

# Thought: The bedroom doesn't have the materials I need. I should go to a place where paint and painting supplies are likely to be found, such as an art studio.
teleport("art studio")
# Observation: You teleport to the art studio.

# Thought: Now that I'm in the art studio, I should look around to find the necessary materials to create green paint.
look_around()
# Observation: This room is called the art studio. In it, you see:
    the agent
    a substance called air
    a large cupboard. The large cupboard door is open. In the large cupboard is: a drawer, a wood cup (containing yellow paint), a wood cup (containing blue paint), a wood cup (containing red paint).
    a table. On the table is: a glass cup (containing nothing).
    a wood cup (containing yellow paint)
    a wood cup (containing blue paint)
    a wood cup (containing red paint)

# Thought: I see blue paint here, which is one of the primary colors needed to mix green paint. I should start by pouring it into an empty container to mix with another color.
pour("cup containing blue paint", "glass cup")
# Observation: You pour the contents of the wood cup into the glass cup.

# Thought: Yellow paint is the other primary color needed to mix with blue to create green. I should add the yellow paint to the container with blue paint.
pour("cup containing yellow paint", "glass cup")
# Observation: You pour the contents of the wood cup into the glass cup.

# Thought: Now that I have both blue and yellow paint in the same container, I need to mix them together to create green paint.
mix("glass cup")
# Observation: Blue and yellow paint mix to produce green paint.

# Thought: The task is to create green paint and then focus on it. Now that the green paint is made. Before I focus on the green paint, I should look at the glass cup to confirm it contains green paint.
look_at("glass cup")
# Observation: The glass cup contains green paint.
```

Your response can be:

```python
# Thought: The task is to create green paint and then focus on it. Now that the green paint is made, I should concentrate on the final product.
focus_on("green paint")
```
\end{promptbox}
\begin{promptbox}{ICL Example for ScienceWorld Batched}
```python
# Task Description: Your task is to use chemistry to create green paint. When you are done, focus on the green paint.

# Thought: I need to find materials to create green paint, which are likely to be found in an art studio.
look_around()
# Observation: This room is called the bedroom. In it, you see:
    the agent
    a substance called air
    a bed. On the bed is: a mattress. On the mattress is: a white pillow..
    a closet. The closet door is open. In the closet is: nothing.
    a picture
    a table. On the table is: nothing.

# Thought: The bedroom doesn't have the materials I need. I should go to a place where paint and painting supplies are likely to be found, such as an art studio.
teleport("art studio")
# Observation: You teleport to the art studio.

# Thought: Now that I'm in the art studio, I should look around to find the necessary materials to create green paint.
look_around()
# Observation: This room is called the art studio. In it, you see:
    the agent
    a substance called air
    a large cupboard. The large cupboard door is open. In the large cupboard is: a drawer, a wood cup (containing yellow paint), a wood cup (containing blue paint), a wood cup (containing red paint).
    a table. On the table is: a glass cup (containing nothing).
    a wood cup (containing yellow paint)
    a wood cup (containing blue paint)
    a wood cup (containing red paint)
```

Your response can be:

```python
# Thought: I see blue paint here, which is one of the primary colors needed to mix green paint. I should start by pouring it into an empty container to mix with another color.
pour("cup containing blue paint", "glass cup")

# Thought: Yellow paint is the other primary color needed to mix with blue to create green. I should add the yellow paint to the container with blue paint.
pour("cup containing yellow paint", "glass cup")

# Thought: Now that I have both blue and yellow paint in the same container, I need to mix them together to create green paint.
mix("glass cup")

# Thought: The task is to create green paint and then focus on it. Now that the green paint is made, I should concentrate on the final product.
focus_on("green paint")
```
\end{promptbox}

\subsection{EB-Habitat}

\subsubsection{Instructions}

\begin{promptbox}{Instruction for EB-Habitat EOCV, Reactive, Batched}
## You are a robot operating in a home. Given a task, you must accomplish the task using a defined set of actions to achieve the desired outcome.

## Action Descriptions and Validity Rules
- Navigation: Parameterized by the name of the receptacle to navigate to. This skill is always valid.
- Pick: Parameterized by the name of the object to pick. Only valid if the robot is close to the object, not holding another object, and the object is not inside a closed receptacle. Note that some object can be out of observation when you are close to it, so always try to pick the object up after navigating to somewhere.
- Place: Parameterized by the name of the receptacle to place the object on. Only valid if the robot is close to the receptacle and is holding an object.
- Open: Parameterized by the name of the receptacle to open. Only valid if the receptacle is closed and the robot is close to the receptacle.
- Close: Parameterized by the name of the receptacle to close. Only valid if the receptacle is open and the robot is close to the receptacle.

## The available action id (0 ~ <action_count>) and action names are: <action_list>.

<icl_example>

## Guidelines
0. **Reflect on yourself**: You should first reflect in one sentence on the agent's state in relation to the task goal, and then output your next actions.
1. **Output Plan**: The task is not done yet, avoid generating empty plan.
2. **Searching for objects**: When searching for some object, navigate around the room, and try to pick up specific object. Note that some object can be out of observation when you are close to it, so always try to pick the object up after navigating to somewhere even if you can't see it in the observation. The robot is using a suction cup robotic arm, so the picked up object will be floating below the arm after being picked up.
3. **Action Validity**: Make sure match the action name and its corresponding action id in the output.\n Avoid performing actions that do not meet the defined validity criteria.
4. **Prevent Repeating Action Sequences**: Do not repeatedly execute the same action or sequence of actions.\n Try to modify the action sequence because previous actions do not lead to success.
5. **Multiple Instances**: There may be multiple instances of the same object, distinguished by an index following their names, e.g., cabinet 2, cabinet 3. You can explore these instances if you do not find the desired object in the current receptacle.
6. **Reflection on History and Feedback**: Use interaction history and feedback from the environment to refine and enhance your current strategies and actions. If the last action is invalid, reflect on the reason, such as not adhering to action rules or missing preliminary actions, and adjust your plan accordingly.

=======Reactive=======
You are required to generate only 1 action for the plan.
======================

========Batched=======
You should generate at most 5 actions in the plan.
======================

=========EOCV=========
You should generate at most 5 actions in the plan. Additionally, in order to verify if your plan is executable, you need to output your expected observation(described in text) after executing each action in the plan. The expected visual state should be concise and only include key information such as object locations and states, without detailed descriptions.
======================

\end{promptbox}

\subsubsection{ICL Examples}

Note that ICL examples in EB-Habitat serves solely to demonstrate the execution flow of the benchmark to the executor; during actual execution, the executors rely on visual observations, and the EOCV executor is forbidden to perform non-information-seeking actions after skipping information-seeking actions.

\begin{promptbox}{ICL Example for EB-Habitat EOCV}
Example 1:
Human instruction: Put both a toy airplane and a bowl onto the black table.
Output: {'language_plan': 'To achieve the goal, the robot must locate the toy airplane and the bowl in the room and then move each item to the black table. There are two tables in the room, and the robot identifies table 1 as the black table. Therefore, the objective is to place both objects on table 1. The plan is as follows: first, navigate to the sofa, pick up the airplane, move to table 1, and place the airplane there. Then, proceed to table 2, where the bowl might be, pick up the bowl, return to table 1, and set the bowl there.', 'executable_plan': [{'action_id': 12, 'action_name': 'navigate to the sofa', 'expected_observation': 'None'}, {'action_id': 47, 'action_name': 'pick up the toy airplane', 'expected_observation': 'The toy airplane is picked up by the robot.'}, {'action_id': 6, 'action_name': 'navigate to the table 1', 'expected_observation': 'None'}, {'action_id': 50, 'action_name': 'place at the table 1', 'expected_observation': 'The toy airplane is placed on the table 1.'}, {'action_id': 7, 'action_name': 'navigate to the table 2', 'expected_observation': 'None'}]}

Example 2:
Human instruction: I made a mistake and left the fridge open. Can you assist me by closing it?
Output: {'language_plan': 'The objective is for the robot to close the refrigerator. To do so, the robot first navigates to the refrigerator and then closes it.', 'executable_plan': [{'action_id': 13, 'action_name': 'navigate to the refrigerator', 'expected_observation': 'None'}, {'action_id': 61, 'action_name': 'close the refrigerator', 'expected_observation': 'The refrigerator is closed.'}]}

\end{promptbox}

\begin{promptbox}{ICL Example for EB-Habitat Reactive/Batched}
Example 1:
Human instruction: Put both a toy airplane and a bowl onto the black table.
Output: {'language_plan': 'To achieve the goal, the robot must locate the toy airplane and the bowl in the room and then move each item to the black table. There are two tables in the room, and the robot identifies table 1 as the black table. Therefore, the objective is to place both objects on table 1. The plan is as follows: first, navigate to the sofa, pick up the airplane, move to table 1, and place the airplane there. Then, proceed to table 2, where the bowl might be, pick up the bowl, return to table 1, and set the bowl there.', 'executable_plan': [{'action_id': 12, 'action_name': 'navigate to the sofa'}, {'action_id': 47, 'action_name': 'pick up the toy airplane'}, {'action_id': 6, 'action_name': 'navigate to the table 1'}, {'action_id': 50, 'action_name': 'place at the table 1'}, {'action_id': 7, 'action_name': 'navigate to the table 2'}]}

Example 2:
Human instruction: I made a mistake and left the fridge open. Can you assist me by closing it?
Output: {'language_plan': 'The objective is for the robot to close the refrigerator. To do so, the robot first navigates to the refrigerator and then closes it.', 'executable_plan': [{'action_id': 13, 'action_name': 'navigate to the refrigerator'}, {'action_id': 61, 'action_name': 'close the refrigerator'}]}

\end{promptbox}

\end{document}